\documentclass[mnsc,nonblindrev]{informs3} 

\OneAndAHalfSpacedXI

\usepackage{multirow}
\usepackage[T1]{fontenc}
\usepackage{makecell}
\usepackage{booktabs}
\usepackage{tablefootnote} 
\usepackage[table,RGB]{xcolor}
\usepackage{soul}

\usepackage{natbib}
 \bibpunct[, ]{(}{)}{,}{a}{}{,}%
 \def\bibfont{\small}%

\newtheorem{definition}{Definition}
\newtheorem{proposition}{Proposition}
\newtheorem{proposition_appendix}{Proposition}
\ECRepeatTheorems

\EquationsNumberedThrough    

\MANUSCRIPTNO{MS-0001-1922.65}

\begin{document}


\RUNAUTHOR{Yang et al.} 

\RUNTITLE{From Machine Learning to Machine Unlearning}

\TITLE{From Machine Learning to Machine Unlearning: Complying with GDPR's Right to be Forgotten while Maintaining Business Value of Predictive Models}

\ARTICLEAUTHORS{%
\AUTHOR{Yuncong Yang$^{\dagger}$}
\AFF{School of Information Management and Engineering, Shanghai University of Finance and Economics, Shanghai 200433, China, \EMAIL{yycphd@stu.sufe.edu.cn}} 
\AUTHOR{Xiao Han$^{\dagger*}$}
\AFF{School of Economics and Management, Beihang University, Beijing 100191, China, \EMAIL{xh\_bh@buaa.edu.cn}}
\AUTHOR{Yidong Chai$^{*}$}
\AFF{School of Management, Hefei University of Technology, Hefei, Anhui 230009, China, \EMAIL{chaiyd@hfut.edu.cn}}
\AUTHOR{Reza Ebrahimi, Rouzbeh Behnia}
\AFF{School of Information Systems and Management, University of South Florida, Tampa, Florida, FL 33620, USA, \EMAIL{ebrahimim@usf.edu, behnia@usf.edu}}
\AUTHOR{Balaji Padmanabhan}
\AFF{Robert H. Smith School of Business, University of Maryland, College Park, Maryland, MD 20742, USA, \EMAIL{bpadmana@umd.edu}}
} 

\ABSTRACT{%
Recent privacy regulations (\emph{e.g.}, GDPR) grant data subjects the `Right to Be Forgotten' (RTBF) and mandate companies to fulfill data erasure requests from data subjects. However, companies encounter great challenges in complying with the RTBF regulations, particularly when asked to erase specific training data from their well-trained predictive models. While researchers have introduced \emph{machine unlearning} methods aimed at fast data erasure, these approaches often overlook maintaining model performance (\emph{e.g.}, accuracy), which can lead to financial losses and non-compliance with RTBF obligations. This work develops a holistic \emph{machine learning-to-unlearning} framework, called Ensemble-based iTerative Information Distillation (ETID), to achieve efficient data erasure while preserving the business value of predictive models. ETID incorporates a new ensemble learning method to build an accurate predictive model that can facilitate handling data erasure requests. ETID also introduces an innovative distillation-based unlearning method tailored to the constructed ensemble model to enable efficient and effective data erasure. Extensive experiments demonstrate that ETID outperforms various state-of-the-art methods and can deliver high-quality unlearned models with efficiency. We also highlight ETID's potential as a crucial tool for fostering a legitimate and thriving market for data and predictive services.
}



\maketitle

%


The data subject shall have the right to obtain from the controller the erasure of personal data concerning him or her without undue delay and the controller shall have the obligation to erase personal data without undue delay under certain circumstances.

-- \textit{Art. 17 GDPR Right to Be Forgotten (RTBF)}
\section{Introduction}
\label{sec:intro}

Predictive analytics uses statistical techniques and machine learning algorithms to analyze patterns in past data and predict future events or trends~\citep{shmueli11}. It has become a powerful and profitable tool to address crucial business and societal challenges, such as personalized product recommendations~\citep{song19}, health monitoring~\citep{yu21}, and financial fraud detection~\citep{xu22}. Typically, companies collect massive data from individuals or organizations to train predictive models and maintain complete control over the acquired data. Recent privacy laws, including the European Union (EU)'s General Data Protection Regulation\footnote{https://gdpr.eu/article-17-right-to-be-forgotten/} (GDPR), the California Consumer Privacy Act\footnote{https://oag.ca.gov/privacy/ccpa\#sectiond} (CCPA) and Canada's Consumer Privacy Protection Act\footnote{https://ised-isde.canada.ca/site/innovation-better-canada/en/consumer-privacy-protection-act} (CPPA), outline `Right to Be Forgotten' (RTBF) regulations. These RTBF regulations empower the data subjects to retract control over their own data and mandate predictive model holders to respond actively to erasure requests from data subjects.

Adhering to the RTBF regulations is crucial for companies deploying predictive models, as non-compliance can lead to substantial fines. For example, in 2020, Google was fined \$8 million by the Swedish Data Protection Authority\footnote{https://edpb.europa.eu/news/national-news/2020/swedish-data-protection-authority-imposes-administrative-fine-google\_en} and \$670K by the Belgian Data Protection Authority\footnote{https://edpb.europa.eu/news/national-news/2020/belgian-dpa-imposes-eu600000-fine-google-belgium-not-respecting-right-be\_en} for violating GDPR principles related to the RTBF. In addition, data subjects are often hesitant to share their data due to concerns over losing control and privacy risks~\citep{wang21,ghose24}. It is reported that, although 83\% of consumers are willing to share their data in exchange for better-personalized services~\citep{accenture18}, 86\% express growing concerns about data privacy~\citep{kpmg21}. Moreover, a majority of Americans (eight-in-ten) feel they barely have control over the data collected about them by companies or the government~\citep{auxier19}. By granting individuals the right to fully erase their shared data, RTBF regulations can significantly mitigate these concerns and thus foster greater trust and willingness to participate in data sharing~\citep{keandsudhir23}.

However, compliance with RTBF is non-trivial for companies using predictive models built from customer data. Simply deleting individual data samples is often inadequate to fully comply with these requests. It is also vital to update the predictive models trained on this data, as research has demonstrated that these models may retain some unique information about specific training samples~\citep{arpit17}. Consequently, effective data erasure in predictive analytics involves not only deleting individual data samples but also eliminating the information retained in the predictive model, making the model as if it had been trained without those samples. A straightforward approach to addressing this issue is to erase the requested data from the training set and retrain the model from scratch (\emph{i.e.}, na\"ive retraining). However, this method is often impractical, particularly when these data erasure requests are frequently made. For instance, since the EU’s RTBF ruling in 2014, Google has received over 2.4 million requests for data erasure from its intelligent search engine services\footnote{https://blog.google/around-the-globe/google-europe/updating-our-right-be-forgotten-transparency-report/}; the UK Biobank, a critical repository of genetic and medical records used in numerous predictive models, has also reported sporadic requests for the data erasure from both its database and associated models~\citep{ginart19}. In such cases, na\"ive retraining is computationally prohibitive and impractical for responding to frequent data erasure requests, particularly when dealing with predictive models trained using extensive datasets~\citep{villaronga18}.

\emph{Machine unlearning} is an emerging paradigm aimed at \textit{efficiently} erasing information about the requested removal data retained in predictive models~\citep{bourtoule21}. It seeks to expedite the data erasure process by partially retraining or adjusting the predictive model to produce an \textit{accurate} unlearned model that still delivers high-quality predictive services. Without resorting na\"ive retraining, machine unlearning expects the unlearned model to generate predictions \textit{consistent} with those of the na\"ive retrained model while remaining distinguishable from the original model, to ensure effective and \textit{verifiable} data erasure. Overall, machine unlearning is suggested to achieve four key requirements: consistency, accuracy, efficiency, and verifiability~\citep{xu23}.


Despite the progress in this area (we present these main ideas in Section~\ref{sec:literature_machine_unlearning}), current machine unlearning methods prioritize the efficiency of data erasure, often neglecting the need for consistency and accuracy in the resulting models. For example, ensemble-based unlearning methods suggest that an ensemble model comprising multiple sub-models can efficiently manage data erasure requests by partially retraining the sub-models associated with the requested data~\citep{bourtoule21}. However, this often requires training sub-models with limited data, which may significantly compromise prediction accuracy~\citep{xu23}. Distillation-based unlearning methods fine-tune a pre-established predictive model using certain reference models that exclude the data requested erasure, aligning the model’s outputs with those of the reference models and thereby facilitating efficient data erasure~\citep{ma22,kurmanji23}. Unfortunately, these methods face challenges in identifying reference models that closely resemble the na\"ive retrained models without fully retraining from scratch. As a result, they rarely produce unlearned models consistent with the na\"ive retrained ones, despite consistency being critical for compliance with privacy regulations.

In contrast, we emphasize that it is crucial to preserve model accuracy and consistency in machine unlearning to maintain service profitability and avoid penalties for non-compliance. To this end, we propose a novel framework that not only efficiently handles data erasure requests but also ensures that unlearned models retain high predictive accuracy and produce results consistent with those of the na\"ive retrained models. To the best of our knowledge, we are among the first in the business research community to address machine unlearning issues to achieve the RTBF within predictive analytics. Notably, our perspective of this here addresses compliance with privacy regulations while maintaining business value. We summarize our key contributions as follows:
\begin{itemize}
    \item We formulate a new holistic machine learning-to-unlearning problem comprising two closely related sub-problems: the predictive model construction and the unlearning request response. The first focuses on building high-performing models that are able to easily unlearn, while the second aims to design an unlearning method meeting the key unlearning requirements including consistency, accuracy, efficiency and verifiability. 
    \item We introduce a novel framework consisting of two innovative methods, reference-oriented ensemble learning (ROEL) and iterative information distillation (TID). The ROEL method trains each sub-model using the majority of the training data, leading to the construction of an accurate ensemble predictive model. This method also generates retrained-alike reference models without incurring additional computational overhead, preparing for efficient and consistent distillation-based unlearning. Besides, the TID method erases the information of samples requested unlearning from the relevant sub-models under the supervision of retrained-alike reference models. It also rectifies the sub-models using the remaining training data to retain predictive accuracy while further boosting efficiency through parallel computing.
    \item Extensive experiments conducted on two datasets demonstrate that our framework can efficiently erase information from predictive models while preserving accuracy and delivering predictions consistent with those of na\"ive retrained models.
    \item Finally, we analyze the implications of our work for information privacy management and offer several managerial insights for the community. For instance, we demonstrate that adopting our framework not only addresses immediate concerns related to data unlearning but also promotes long-term benefits by fostering a more trustworthy data ecosystem.
\end{itemize}

\section{Problem Formulation and Technical Foundations}
In this section, we formally define the holistic machine learning-to-unlearning problem and present some critical technical foundations. The notations used in this work are summarized in Table~\ref{tab:notation}. 

\begin{table*}\footnotesize
\centering
\caption{Summary of notations.}
\def\arraystretch{1.3}\begin{tabular}{cl}
\toprule
  \textbf{Notations}       & \textbf{Description}   \\  \midrule
 $\mathbf D$, $\mathbf D^u$, $\mathbf D^r$, $\mathbf D^t$   & Training data, Unlearning data, Remaining data, Testing data.   \\  
 $\mathbf X$, $\mathbf X^u$, $\mathbf X^r$, $\mathbf X^t$   & Features of training, unlearning, remaining, testing samples.   \\
 $\mathbf d_i$     &   The $i$-th training data part.   \\
 $\mathbf D_{-i}$   & The $i$-th training data subset including data parts $\{\mathbf d_1,\cdots,\mathbf d_{i-1},\mathbf d_{i+1},\cdots,\mathbf d_K\}$. \\
 $\mathbf d^u_i$     &   Unlearning data in $\mathbf d_i$.   \\
 $M$  & The (ensemble) predictive model trained with training data $\mathbf D$.  \\ 
 $M_i$  & The $i$-th target sub-model trained with the subset $\mathbf D_{-i}$.  \\ 
 $\mathbb M^{tg}$/$M^{tg}_{i}$  & The target sub-model collection/The $i$-th target sub-model.  \\ 
 $M^u_{i}$  & The $i$-th unlearned sub-model.  \\ 
 $M^{rt}_{i}$  & The $i$-th na\"ive retrained sub-model.  \\ 
 \bottomrule
\end{tabular}%
\label{tab:notation}
\end{table*}%

\subsection{Problem Formulation}
\begin{figure}[t]
\begin{center}
\includegraphics[width=0.95\linewidth]{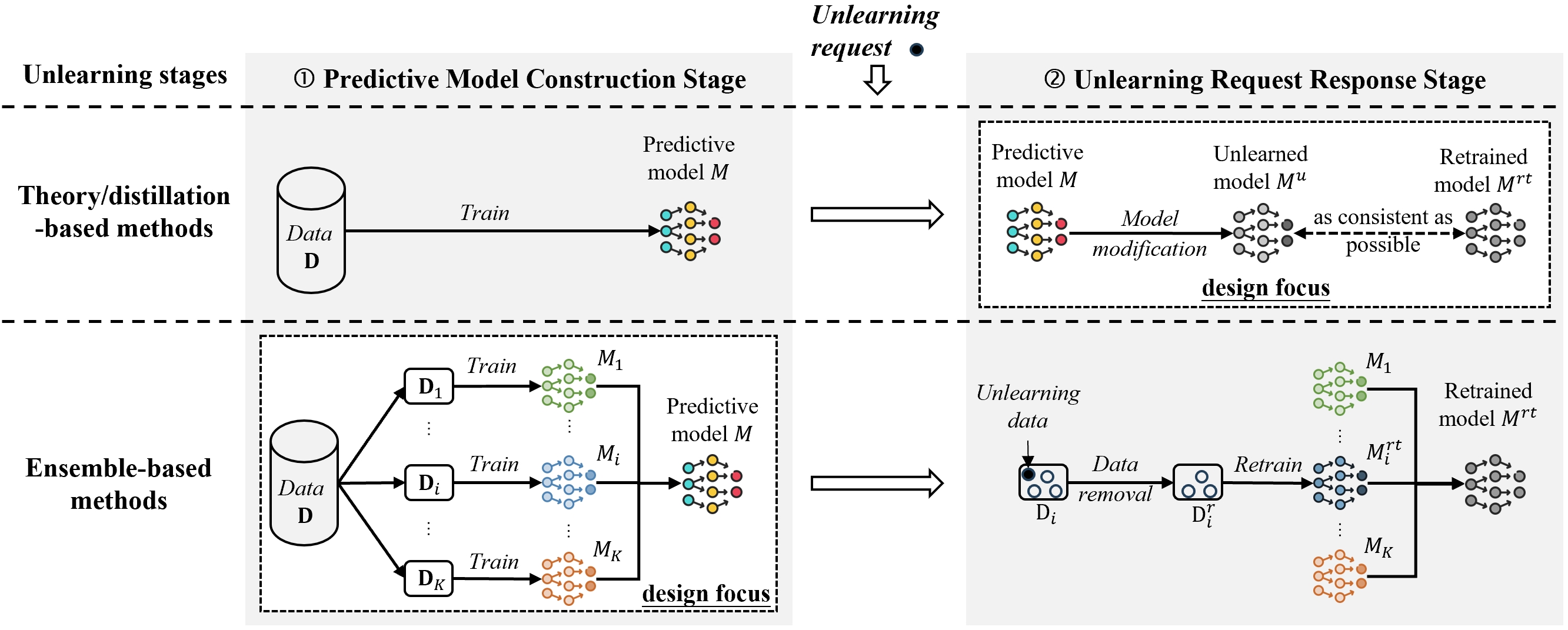}
\caption{The diagram of the general machine unlearning process consists of two stages. Most existing unlearning methods design the unlearning request response stage, while some design the predictive modeling stage.} \label{fig:general_framework}
\end{center}
\end{figure}

The machine unlearning process typically has two distinct stages: the predictive model construction and the unlearning request response. Figure~\ref{fig:general_framework} depicts the general machine unlearning process and illustrates the typical designs of prior methods at each stage. Below, we outline the key concepts involved in these stages and formally define our problem.

\subsubsection{The Predictive Model Construction Stage.} 
During this stage, providers collect data from various data subjects and train a satisfactory predictive model to offer predictive services. Let $\mathbf D \subset \mathbb D$ denote the training dataset, where $\mathbb D$ represents the overall data space. Here, $\mathbf X$ and $\mathbf Y$ denote the features and labels of samples in $\mathbf{D}$, respectively. In practice, providers may adopt arbitrary machine learning algorithms (\emph{e.g.}, neural networks, random forests, or linear models) and produce either a single or an ensemble predictive model. Let $M$ represent the predictive model trained on the dataset $\mathbf D$ using a specific algorithm~$A$. The predictive model $M$ is often referred to as the \textit{target model} for unlearning purposes as well.

\subsubsection{The Unlearning Request Response Stage.}
Once the predictive service is in use, providers may receive requests from data subjects to erase specific data samples. The data requested for erasure, referred to as \textit{unlearning data} and denoted by $\mathbf D^{u}$, may consist of one or more samples from the training data $\mathbf D$\footnote{It is a common assumption in machine unlearning research that all the potential unlearning samples come from the training data $\mathbf{D}$~\citep{caoandyang15, bourtoule21}.}. To comply with the RTBF regulations, providers must ensure that $\mathbf{D}^{u}$ is thoroughly erased from the trained models, as these models may retain unique information of the training samples~\citep{caoandyang15,arpit17}. The most straightforward approach is to retrain the model from scratch using the remaining data $\mathbf D^r=\mathbf D / \mathbf D^u$; however, this approach becomes computationally expensive, particularly with large datasets or complex models, making it impractical for frequent unlearning requests~\citep{bourtoule21}. Consequently, machine unlearning has emerged as an alternative to na\"ive retraining for erasing data's information from a predictive model, with the goal of achieving the following desiderata:
\begin{definition}[The Desiderata of Machine Unlearning] \label{def:Unlearning}
Given a target model $M$ and unlearning data $\mathbf D^u \subset \mathbf D$, machine unlearning seeks to avoid na\"ive retraining and aims to generate an unlearned model $M^u$ that meets the following desiderata~\citep{bourtoule21,xu23}:
\begin{itemize}
    \item \textbf{Consistency}: machine unlearning requires the predictions of $M^u$ for any sample closely resemble those of the na\"ive retrained model $M^{rt}$. While the na\"ive retrained model $M^{rt}$ represents the ideal outcome of a machine unlearning method, the consistency in predictions between $M^u$ and $M^{rt}$ indicates the effectiveness of the unlearning method. The closer the predictions of $M^u$ align with those of $M^{rt}$, the more effective the machine unlearning method is.
    \item \textbf{Accuracy}: it demands that $M^u$ delivers accurate predictions for the remaining samples $\mathbf D^r$, the testing samples $\mathbf D^t$, and the unlearning samples $\mathbf D^u$. A preferable unlearning method should preserve prediction performance so that its unlearned model continues to offer high-quality predictions, which are valuable from a practical business perspective.
    \item \textbf{Efficiency}: it desires a prompt unlearning process with low computational costs. Machine unlearning methods are expected to incur significantly lower computational costs compared to the na\"ive retraining method. 
    \item \textbf{Verifiability}: it stipulates that $M^u$ should be sufficiently distinguishable from $M$ through a verification method, thereby ensuring the discernibility of the unlearning process. Prior literature typically performs verification using membership inference approaches, which determine whether a sample is a member of the predictive model's training data~\citep{ma22}. In principle, the unlearning samples are members of data used to train the original target model but presumably are not members of the unlearned model if they are successfully erased. Therefore, a machine unlearning method is considered verifiable if there is a significant difference in membership inference results for the unlearning data between the original target and unlearned models.
\end{itemize}
\end{definition}

\subsubsection{Problem Definition.}
Current research typically approaches machine unlearning as a single-stage problem, which can be categorized into two streams. The first stream assumes that providers have already established a predictive model and focuses on the unlearning request response stage by carefully modifying the existing model to erase unlearning data~\citep{guo20,kurmanji23}. Unfortunately, these predictive models are generally designed for high performance rather than ease of unlearning, making the unlearning process challenging and inefficient. The second stream mainly addresses the unlearning issue during the model construction stage. This stream of approaches aims to develop models capable of partial retraining when unlearning requests arise, allowing for selective retraining of model parts associated with the unlearning data~\citep{bourtoule21}. Nevertheless, without careful design of the unlearning response method, even partial retraining can still be time-consuming, especially when the unlearning data affects large portions of the model, necessitating extensive retraining. Additionally, the second stream does not sufficiently consider the model's predictive performance in its design.

Unlike existing studies, we approach machine unlearning as a holistic problem encompassing both predictive model construction and unlearning request response. We highlight that predictive service providers can proactively account for unlearning needs during the model construction stage. Constructing the predictive model with foresight is essential for maintaining service performance and streamlining the unlearning response. Additionally, even with models specifically designed for easy unlearning, it remains crucial to tailor the unlearning response methods to guarantee the effectiveness and efficiency of the unlearning process. Below, we formally define our problem:
\begin{definition}[The Holistic Machine Learning-to-Unlearning Problem] 
\label{def:holistic_unlearning} Given a training dataset $\mathbf D$, the holistic machine learning-to-unlearning problem aims to construct an accurate predictive model that can efficiently and effectively erase the information of arbitrary given unlearning data $\mathbf D^u \subset \mathbf{D}$ from the predictive model when requested. Specifically, this problem can be decomposed into two sub-problems:
\begin{itemize}
    \item \textbf{Sub-problem 1 - predictive model construction}: During the predictive model construction stage, given a training dataset $\mathbf D$, design a predictive model $M$ that delivers strong predictive performance while enabling easy erasure of any training data when necessary;
    \item \textbf{Sub-problem 2 - unlearning request response}: During the unlearning request response stage, given the previously constructed model $M$ and the unlearning data $\mathbf D^u$, design an unlearning response method that meets all machine unlearning desiderata in Definition~\ref{def:Unlearning}.
\end{itemize}
\end{definition}

\textbf{Remark}. Achieving all desiderata of machine unlearning is exceedingly challenging and sometimes impractical in certain scenarios. For instance, when handling a large amount of unlearning data, ensuring a consistent unlearned model may inevitably lead to its inaccuracy because the unlearned model is restricted to preserving knowledge derived from a smaller subset of the original training data. Notably, existing machine unlearning research has predominantly focused on meeting the requirements of \textit{verifiability} and \textit{efficiency}, with less emphasis on \textit{accuracy} and \textit{consistency}~\citep{guo20,kurmanji23}. However, from a business perspective, these two requirements are critical since \textit{accuracy} directly reflects the quality of predictive services, which produces profitability, and \textit{consistency} relates to the effectiveness of unlearning and is crucial for ensuring compliance with privacy regulations, thereby helping providers avoid fines. As a pioneering business-oriented study in machine unlearning, we emphasize the importance of accuracy and consistency. Our goal is to develop an innovative machine unlearning method that not only enhances accuracy and consistency but also maintains efficiency and verifiability.

\subsection{Technical Foundations}
In this section, we introduce some preliminaries of two crucial techniques commonly employed in previous unlearning studies, which will also form the basis of our design. They are ensemble learning and distillation learning.

\subsubsection{Ensemble Learning.}
Ensemble learning is a widely used machine learning technique that aims to achieve high-performing predictive models by combining multiple sub-models~\citep{breiman96}. A common ensemble approach typically partitions a training dataset $\mathbf D$ into several subsets, with each subset used to train a separate sub-model $M_i$. The results of these sub-models are then aggregated to produce the final output of the ensemble predictive model $M$.

Prior unlearning studies reveal the potential of ensemble learning for constructing models that are able to easily unlearn~\citep{bourtoule21}. Essentially, with ensemble-based methods, handling an unlearning request only requires partially retraining one or a few of these sub-models, making the retraining process more efficient. While these studies emphasize the ease of unlearning by partial retraining, they unfortunately neglect the quality of the model, which may lead to inaccurate prediction performance.


\subsubsection{Distillation Learning.} 
\label{sec:distillation}
Distillation learning is a technique that fine-tunes a predictive model $M$ under the supervision of a reference model $M^{rf}$, aiming to align $M$'s predictions on a set of distillation data $\mathbf{D}_{dis}$ with those of $M^{rf}$~\citep{hinton15}. This technique is commonly used in model compression and knowledge transfer, where a well-trained large model serves as the reference to guide the training of a small, randomly initialized model. Specifically, the posterior probabilities produced by the reference model $M^{rf}$ are used as "soft targets" for training the small model $M$. By ensuring that $M$'s predictions are consistent with these soft targets, the knowledge acquired by the reference model $M^{rf}$ can be effectively transferred to $M$. Let $\mathbf X_{dis}$ denote the features of distillation data. The objective of distillation learning is generally formulated as:
\begin{equation}
    \min_{M} \textsc{diff}(M^{rf}(\mathbf X_{dis}),M(\mathbf X_{dis}))
\end{equation}
where $\textsc{diff}(\cdot)$ quantifies the difference between two models' predictions.

While retraining a model from scratch is computationally expensive, fine-tuning is typically more efficient, as providers only need to slightly adjust the parameters with fewer data instead of training the entire model from the ground up. Therefore, recent seminal works delve into distillation learning techniques for machine unlearning \citep{ma22,kurmanji23}. Ideally, distillation can assist in unlearning a target model $M$ by fine-tuning $M$ under the supervision of its na\"ive retrained model $M^{rt}$ as a reference model (\emph{i.e.,} $M^{rf}=M^{rt}$). Through fine-tuning $M$, the goal is to obtain an unlearned model $M^u$ whose predictions align with those of $M^{rt}$ for arbitrary data $\mathbf{\Tilde{D}} \subset \mathbb D$ with features $\mathbf{\Tilde{X}}$. In general, the information of the unlearning data can be considered completely erased from the target model if its predictions on arbitrary data are consistent with those of the na\"ive retrained model. This can be achieved by solving the following optimization problem:
\begin{equation}
\label{eq:distllation}
    M^u = \min_{M} \textsc{diff}(M^{rt}(\mathbf{\Tilde{X}}),M(\mathbf{\Tilde{X}}))
\end{equation}
Nevertheless, without retraining the target model from scratch, its na\"ive retrained model $M^{rt}$ is not accessible as a distillation reference. Although previous distillation-based machine unlearning methods are highly efficient, they fail to identify an appropriate reference model, resulting in inaccurate and inconsistent unlearning models.

\section{Literature Review}
\label{sec:leterature}
Our study closely relates to two main research streams: information privacy management and machine unlearning. We review each of these streams and emphasize the key contributions and innovations of our work in this section.


\subsection{Information Privacy Management}
Privacy management has been a key area of focus in Information Systems (IS) research for decades~\citep{smith96,smith11,pavlou11,cichy21}. Recently, the widespread adoption of artificial intelligence (AI) and machine learning has introduced both opportunities through predictive modeling and novel challenges for information privacy management. In response, a substantial body of IS research has emerged to address these challenges, which can be categorized into two primary streams~\citep{xuanddinev22}: empirical studies aimed at understanding consumers' privacy concerns and behaviors~\citep{acquisti15,acquisti16,xuandzhang22b}, and technical studies focused on developing privacy assurance techniques to protect sensitive information from privacy breaches~\citep{liandqin17,han21}.

Empirical studies on information privacy management focus on understanding the factors driving people's privacy concerns and behaviors~\citep{awadandkrishnan06,acquisti15,xuandzhang22a}. For instance, \citet{sutanto13} explored the personalization-privacy paradox in the context of personalized smartphone advertising, examining how privacy impacts the process and content gratifications derived from personalization and how IT solutions can be designed to alleviate privacy concerns. While most studies considered only one or a few specific contexts,~\cite{xuandzhang22a} developed a conceptual and quantitative framework to examin the multiplicity of contexts and their impact on consumers' cognition and perceptions of privacy. Additionally, some recent empirical work has studied the impacts of privacy laws and regulations on companies and individuals~\citep{johnson23,keandsudhir23}. Specifically, ~\cite{keandsudhir23} studied GDPR’s equilibrium impact using a dynamic two-period model of forward-looking companies and consumers. They found that privacy rights would reduce consumers’ hold-up concerns and raise (reduce) firm profit and social welfare when privacy breach risk is high (low). 
Compared to these empirical studies, which focus on understanding or explaining privacy-related behaviors, we highlight that developing privacy-respecting AI technologies is one of the key managerial issues for IS researchers as well~\citep{hevner04,berente21}. In this research, we aim to design a novel machine unlearning method as a practical privacy assurance technique to efficiently and effectively protect the RTBF for data subjects in predictive analytics.

Technical research aims at designing privacy assurance techniques to shield personal private information from unauthorized access or inference through public data sources~\citep{xuanddinev22}. In the early stages of IS development, privacy concerns centered around data flows, known as \textit{data-centered privacy concerns}. For example, earlier studies demonstrated that consumers' private information (\emph{e.g.}, identities) could be acquired by linking two publicly available datasets~\citep{sweeney97}. To avoid such personal information leakage from unauthorized access, scholars developed techniques that "anonymize" datasets before data sharing by preventing any record from being linked to an individual while retaining the data value for analytic purposes~\citep{liandsarkar13,liandsarkar14,menonandsarkar16,liandqin17,li23}. While these studies investigated privacy-preserving data sharing through anonymization, recent work has proposed leveraging secure multiparty computation technique in financial network analytics. This approach addresses data-centered privacy concerns by enabling institutions to locally compute on their real data without sharing it~\citep{hastings23}. With the rise of AI technologies, many organizations and companies started to use predictive analytic tools to unlock the value of their collected data. Nevertheless, despite the benefits, it has introduced new attack surfaces and shifted privacy concerns from data-centered to \textit{knowledge-centered}~\citep{xuanddinev22}. These privacy concerns concentrate on the private knowledge that could be inferred from the collected data. For instance, the private information (\emph{e.g.}, location) of users in online social networks can be precisely inferred based on their publicly available data, even though they intentionally hide this information~\citep{han21}. As data anonymization methods have proven ineffective in preventing the inference of private knowledge, the vast majority of privacy assurance studies turned to other techniques, such as suppression and differential privacy~\citep{dwork06}. Suppression techniques selectively hide or obfuscate portions of users' publicly available data, thereby preventing their private knowledge from being inferred by adversaries~\citep{han21,macha24}. Differential privacy, on the other hand, does not prevent adversaries from gaining insights but ensures that any knowledge obtained could have been inferred even without access to an individual's specific data~\citep{chenprivacy22,lei23}. While these privacy assurance techniques effectively thwart specific privacy attacks, their main objective is to prevent the acquisition or inference of private information. In contrast, our focus is to develop a technique that completely erases the information of requested data from predictive models to comply with the RTBF requirement. Table~\ref{tab:RelatedWork1} summarizes the recent technical research on information privacy management for comparison.
\definecolor{mycolor}{RGB}{192, 192, 192}
\definecolor{black}{RGB}{0, 0, 0}
\begin{table*}[t]\scriptsize
\centering
\caption{Selected recent technical research on information privacy management.}
\def\arraystretch{1.3}\begin{tabular}{@{}cllll@{}}
\toprule
  \textbf{Year}  & \textbf{Author(s)}  &  \textbf{Research Focus}  & \textbf{Privacy Issue} & \textbf{Technique}   \\   \midrule   \arrayrulecolor{mycolor}
  2013  & Li and Sarkar  &  Privacy-preserving data sharing and analytics   & \begin{tabular}[l]{@{}l@{}}Data-centered \\privacy acquisition \end{tabular} & Anonymization   \\   \cmidrule{1-5}  \arrayrulecolor{mycolor}
  2014  & Li and Sarkar  &  Protecting data privacy from regression attacks   & \begin{tabular}[l]{@{}l@{}}Data-centered \\privacy acquisition \end{tabular} & Anonymization   \\   \cmidrule{1-5}  \arrayrulecolor{mycolor}
  2016  & Menon and Sarkar  &  Sanitizing large transactional databases for sharing  & \begin{tabular}[l]{@{}l@{}}Data-centered \\privacy acquisition \end{tabular} & Anonymization   \\   \cmidrule{1-5}  \arrayrulecolor{mycolor}
  2017  & Li and Qin  &  Anonymizing unstructured text data for sharing & \begin{tabular}[l]{@{}l@{}}Data-centered \\privacy acquisition \end{tabular} & Anonymization   \\   \cmidrule{1-5}  \arrayrulecolor{mycolor}
  2021  & Han et al.  &  \begin{tabular}[l]{@{}l@{}}Estimating and managing the exposure risk of users' \\hidden information in online social networks \end{tabular} & \begin{tabular}[l]{@{}l@{}}Knowledge-centered \\privacy inference \end{tabular} & Suppression   \\   \cmidrule{1-5}   \arrayrulecolor{mycolor}
  2022  & Chen et al.  &  \begin{tabular}[l]{@{}l@{}}Privacy-preserving dynamic personalized pricing in \\online learning settings \end{tabular} & \begin{tabular}[l]{@{}l@{}}Knowledge-centered \\privacy inference \end{tabular} & Differential privacy   \\   \cmidrule{1-5}  \arrayrulecolor{mycolor}
  2023  & Hastings et al.  &  \begin{tabular}[l]{@{}l@{}}Privacy-preserving network analytics by computing \\privately on the real data held by the data holders \end{tabular} & \begin{tabular}[l]{@{}l@{}}Data-centered \\privacy acquisition \end{tabular} & \begin{tabular}[l]{@{}l@{}}Secure multiparty \\computation \end{tabular}  \\   \cmidrule{1-5}  \arrayrulecolor{mycolor}
  2023  & Lei et al.  &  Privacy-preserving personalized offline pricing  & \begin{tabular}[l]{@{}l@{}}Knowledge-centered \\privacy inference \end{tabular} & Differential privacy   \\   \cmidrule{1-5}  \arrayrulecolor{mycolor}
  2023  & Li et al.  &  \begin{tabular}[l]{@{}l@{}}Protecting privacy from re-identification risks in \\panel data \end{tabular} & \begin{tabular}[l]{@{}l@{}}Data-centered \\privacy acquisition \end{tabular} & Anonymization   \\   \cmidrule{1-5}  \arrayrulecolor{black}
  2024  & Macha et al.  &  \begin{tabular}[l]{@{}l@{}}Quantifying and reducing personalized privacy risks \\in consumer mobile trajectories sharing \end{tabular}  & \begin{tabular}[l]{@{}l@{}}Knowledge-centered \\privacy inference \end{tabular} & Suppression   \\   \midrule 
  2024  & Ours  &  \begin{tabular}[l]{@{}l@{}}Erasing requesting data from predictive models \\efficiently and effectively to protect the RTBF \end{tabular} & \begin{tabular}[l]{@{}l@{}}RTBF: private data \\retraction \end{tabular} & Machine unlearning \\
\bottomrule  
\end{tabular}%
\label{tab:RelatedWork1}
\end{table*}

\subsection{Machine Unlearning Studies}
\label{sec:literature_machine_unlearning}
Since the implementation of the RTBF regulation ruling by the highest court in the EU in 2014, there has been a lot of pioneering research in the field of machine unlearning~\citep{caoandyang15,xu23}. Specifically, existing machine unlearning methods can be categorized into three types: ensemble-based, theory-based, and distillation-based methods. Typically, ensemble-based methods design the predictive model construction stage to develop an ensemble model that is inherently easy to retrain and respond to the following unlearning requests by simply retraining parts of the model. Comparatively, theory-based and distillation-based methods focus on the unlearning request response stage and design various post-model modification approaches on pre-established predictive models to address data erasure requests.

Ensemble-based unlearning methods aim to reduce the computational costs of retraining by only retraining parts of the trained predictive model. To achieve this goal, they carefully designed the predictive model construction method to establish an ensemble predictive model. In particular, \citet{bourtoule21} proposed the \textit{Sharded, Isolated, Sliced and Aggregated} (\emph{i.e.}, SISA), which first randomly splits the training set into several non-overlapping subsets, and trains a sub-model on each subset separately. After that, the final prediction results are obtained from the aggregation of sub-models through majority voting or averaging. During the unlearning response stage, only the sub-models trained using unlearning samples are retrained. Based on the same core idea, RecEraser \citep{chenrec22} and GraphEraser \citep{chengraph22} introduce ensemble-based unlearning methods to deal with unlearning requests in recommender systems and graphs, respectively. However, the ensemble-based methods may still face inefficiency problems when unlearning samples are spread across various subsets since they need to retrain numerous sub-models. Besides, partitioning the training dataset into too many disjoint subsets may result in sub-models being trained with insufficient samples, thereby leading to a decrease in model accuracy performance. Our framework also employs ensemble modeling to build predictive models, but it significantly differs in design from these methods. On the one hand, it ensures that each data subset contains the majority of the training samples, thus preventing the issue of diminishing the predictive accuracy due to undertrained sub-models. On the other hand, the modeling process is not aimed at facilitating retraining but is designed to efficiently and consistently accomplish the distillation-based unlearning process.

Theory-based methods design post-model modification approaches that modify the parameters of a pre-established predictive model premised in some theory. One stream of them precisely calculates the information of unlearning data retained in the target model; it then utilizes the differential privacy theory~\citep{dwork06} to add noise on the parameters of the target model for generating an unlearned model that resembles the na\"ive retrained model~\citep{ginart19, guo20,izzo21}. Nevertheless, these methods are only limited to simple algorithms, \emph{e.g.}, K-Means, or linear models. Another stream incorporates information theory such as Fisher and Shannon mutual information to establish model-independent data information estimation for machine unlearning. They employ Hessian or linearization approximations for complex models, and erase information of the unlearning data based on the approximated model parameters~\citep{golatkar20a,golatkar20b,golatkar21}. However, these methods often introduce inconsistent unlearning results because of model approximations, and the information theory-based computation process remains costly. 


Distillation-based methods fine-tune a pre-established predictive model by making its outputs of the data samples required to be forgotten aligned with some alternative references. Specifically, Relabel \citep{graves21} assigns new random labels to the unlearning data as their references. On the other hand, Forsaken \citep{ma22} uses outputs from some testing samples as references, distilling the unlearning samples by enforcing the target model to produce output distributions on these samples similar to those on the testing samples. Unlike Forsaken, which uses external testing sample outputs as references, recent works suggest generating references based on the unlearning samples themselves. For instance, Bad-T \citep{chundawat23} employs a stochastically initialized model to produce random outputs for unlearning samples, guiding the target model to also generate random outputs. However, this approach can significantly reduce model accuracy as the stochastically initialized model diverges from a na\"ive retrained model. SCRUB \citep{kurmanji23} uses the original trained model's outputs on unlearning samples as references, aiming to make the target model's outputs on these samples as dissimilar to the references as possible. AFS \citep{zhou23} introduces outputs with minimal membership leakage risks as references, adjusting the target model weights using an adversarial membership inference attack (\emph{i.e.}, MIA) module, which helps the target model produce outputs on unlearning samples that are resistant to MIAs. Although membership inference is commonly used as a verification mechanism for machine unlearning, resistance to MIAs does not necessarily imply that samples have been fully unlearned. Overall, existing distillation-based methods often employ unsuitable references that disrupt the true relationships between inputs and labels and fail to mimic the na\"ive retrained model, thereby leading to inaccurate and inconsistent unlearned models.

\begin{table*}[t]\footnotesize
\centering
\caption{Comparison between our framework and existing machine unlearning methods.}
\def\arraystretch{1.3}\begin{tabular}{@{}p{0.25\textwidth}cccccc@{}}
\toprule
  \multirow{3.5}{*}{\textbf{Methods}}  & \multicolumn{2}{c}{\textbf{Design stage}}  & \multicolumn{4}{c}{\textbf{Unlearning desiderata}} \\  \cmidrule(r){2-3} \cmidrule(r){4-7} 
   &    \begin{tabular}[c]{@{}c@{}} \textbf{Model} \\ \textbf{construction} \end{tabular}& \begin{tabular}[c]{@{}c@{}} \textbf{Unlearning} \\ \textbf{response} \end{tabular} & \textbf{Consistency}     & \textbf{Accuracy}     & \textbf{Efficiency}    &  \textbf{Verifiability}   \\  \midrule 
 \begin{tabular}[l]{@{}l@{}} Ensemble-based methods,\\ \emph{e.g.}, \cite{bourtoule21} \end{tabular} &$\checkmark$ &$\times$  &High    & {Low}   & Low     & High    \\  \midrule
 \begin{tabular}[l]{@{}l@{}} Distillation-based methods,\\ \emph{e.g.}, \cite{kurmanji23} \end{tabular}   &$\times$ &$\checkmark$ & Low   & {Low}  & High     & High     \\ \hline
 \begin{tabular}[l]{@{}l@{}} Theory-based methods,\\ \emph{e.g.}, \cite{golatkar20a} \end{tabular}  &$\times$ &$\checkmark$ &Low   & Low  & Low     & High     \\ \midrule
Ours  &  $\checkmark$ &$\checkmark$ & High  & High  & High     & High     \\  \bottomrule
\end{tabular}%
\label{tab:RelatedWork}
\end{table*}%

\subsection{Key Novelty of Our Study}
Although privacy management has been a longstanding focus in business research community, to the best of our knowledge, we are among the first to address issue of machine unlearning for compliance with the RTBF regulations in predictive analytics. While there has been some research on machine unlearning, our literature review underscores several research gaps between existing approaches and our framework, as summarized in Table~\ref{tab:RelatedWork}. A notable distinction is that, in addition to meeting the efficiency and verifiability goals of machine unlearning, we achieve high accuracy and consistency to preserve the quality of predictive services and comply with privacy regulations, thereby maintaining business value. In contrast, existing studies have not adequately considered these aspects, potentially resulting in profit losses and leading to non-compliance with the RTBF regulations. Furthermore, to satisfy all the desiderata of machine unlearning, we propose a holistic machine learning-to-unlearning framework, which integrates a novel model construction method and an innovative unlearning method. Specifically, we introduce a novel ensemble learning method that not only builds a highly accurate predictive model but also provides reference models that closely resemble na\"ive retrained models for the subsequent unlearning. Besides, we design a new distillation-based unlearning method specifically tailored to the established predictive model, enabling efficient unlearning while ensuring the verifiability of the results. Additionally, we leverage the remaining data to rectify the unlearned model, enhancing its overall accuracy.

\section{Proposed Framework}
\label{sec:Framework}
\begin{figure}[t]
\begin{center}
\includegraphics[width=0.8\linewidth]{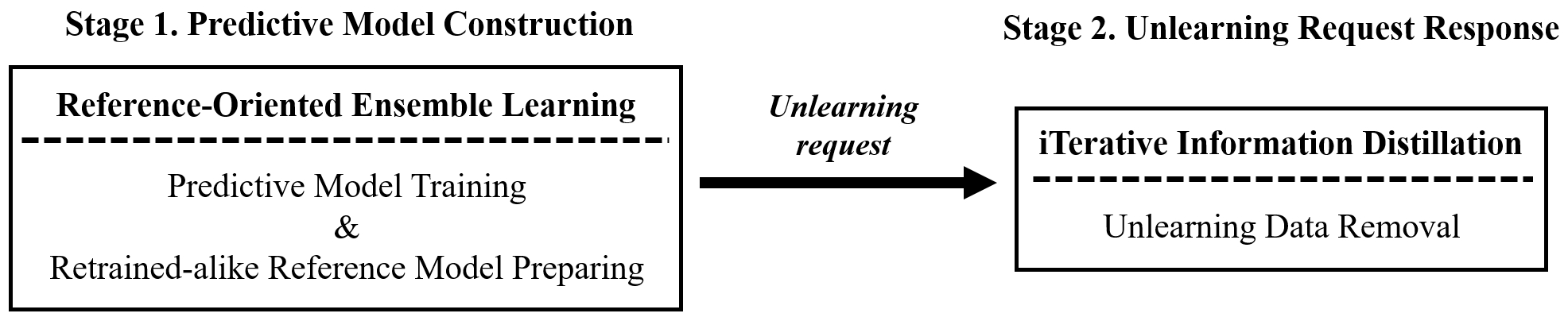}
\caption{Overview of Ensemble-based iTerative Information Distillation (ETID) framework.} \label{fig:framework}
\end{center}
\end{figure}

In this section, we propose a holistic machine learning-to-unlearning framework named \textit{E}nsemble-based i\textit{T}erative \textit{I}nformation \textit{D}istillation (ETID). Figure~\ref{fig:framework} presents an overview of ETID, which introduces two novel methods to address the predictive model construction and unlearning request response sub-problems. Particularly, in the first stage, we introduce a new \textit{R}eference-\textit{O}riented \textit{E}nsemble \textit{L}earning (ROEL) method to train an accurate ensemble predictive model and create retrained-alike models as reference models to facilitate subsequent distillation-based unlearning. In the second stage, we propose an innovative distillation-based unlearning method called i\textit{T}erative \textit{I}nformation \textit{D}istillation (TID) to address unlearning requests, which is tailored to our developed predictive model. Overall, ETID is meticulously designed to improve the consistency and accuracy of machine unlearning while maintaining efficiency and verifiability. The following sections elaborate on these two novel methods within ETID.

\subsection{Reference-Oriented Ensemble Leaning}
\label{sec:ROEL}
 
Before addressing the predictive model construction sub-problem, we first determine an appropriate unlearning strategy to ensure the constructed predictive model can effectively support future unlearning responses. Based on the comprehensive analysis of the strengths and limitations of existing unlearning strategies discussed in Section~\ref{sec:distillation} and Section~\ref{sec:literature_machine_unlearning}, we opt for distillation techniques over partial retraining and theory-based modifications. This is because of the ability of distillation methods to quickly erase data from large models trained on extensive datasets through fine-tuning and their adaptability across various machine learning models. However, a key challenge in developing effective distillation-based unlearning methods is the lack of suitable reference models closely resembling the na\"ive retrained model.

Consequently, in this section, we design a reference-oriented ensemble learning (ROEL) method to construct predictive models with high performance while easing the challenge of distillation-based unlearning. More specifically, by exploring the advantageous properties of ensemble learning, ROEL is designed to achieve a superior predictive model while generating retrained-alike reference models without requiring additional computational resources for future unlearning requests. Typically, an ensemble model comprising well-trained sub-models can attain satisfactory predictive performance~\citep{breiman96}; thus, unlike previous methods that train sub-models on small subsets of samples~\citep{bourtoule21}, ROEL ensures its sub-models are sufficiently trained with the majority of training samples. Furthermore, ROEL is well-structured so that its sub-models can mutually serve as retrained-alike reference models, which are critical for effective and efficient unlearning in the subsequent stage. Below we first define $\delta$-alike and retrained-alike model, and then illustrate the design of ROEL.

\begin{definition}[$\delta$-alike model]
\label{def_delta_alike}
Model $M'=A(\mathbf D')$ is a $\delta$-alike model of model $M=A(\mathbf D)$ if they use the same algorithm $A$ and their shared training samples are $\delta\in [0,+\infty)$ times more than their unique samples:
$$\frac{|\mathbf D \bigcap \mathbf D'|}{\max(|\mathbf D|-|\mathbf D \bigcap \mathbf D'|,|\mathbf D'|-|\mathbf D \bigcap \mathbf D'|)}\geq \delta.$$
\end{definition}
Model $M$ and $M'$ are identical when $\delta \to \infty$; they are $1$-alike if their shared training samples are one time more than the unique samples in the larger dataset between $D$ and $D'$.

\begin{definition}[Retrained-alike model]
\label{def_retrained_alike}
Given a model $M=A(\mathbf D)$, unlearning data $\mathbf D^u \subset \mathbf D$, and the na\"ive retrained model $M^{rt}=A(\mathbf D / \mathbf D^u)$, any $\delta$-alike ($\delta \geq 1$) model of $M^{rt}$ that has not been trained on $\mathbf D^u$ is a retrained-alike model of $M$.
\end{definition}

\begin{figure}[t]
\begin{center}
\includegraphics[width=0.57\linewidth]{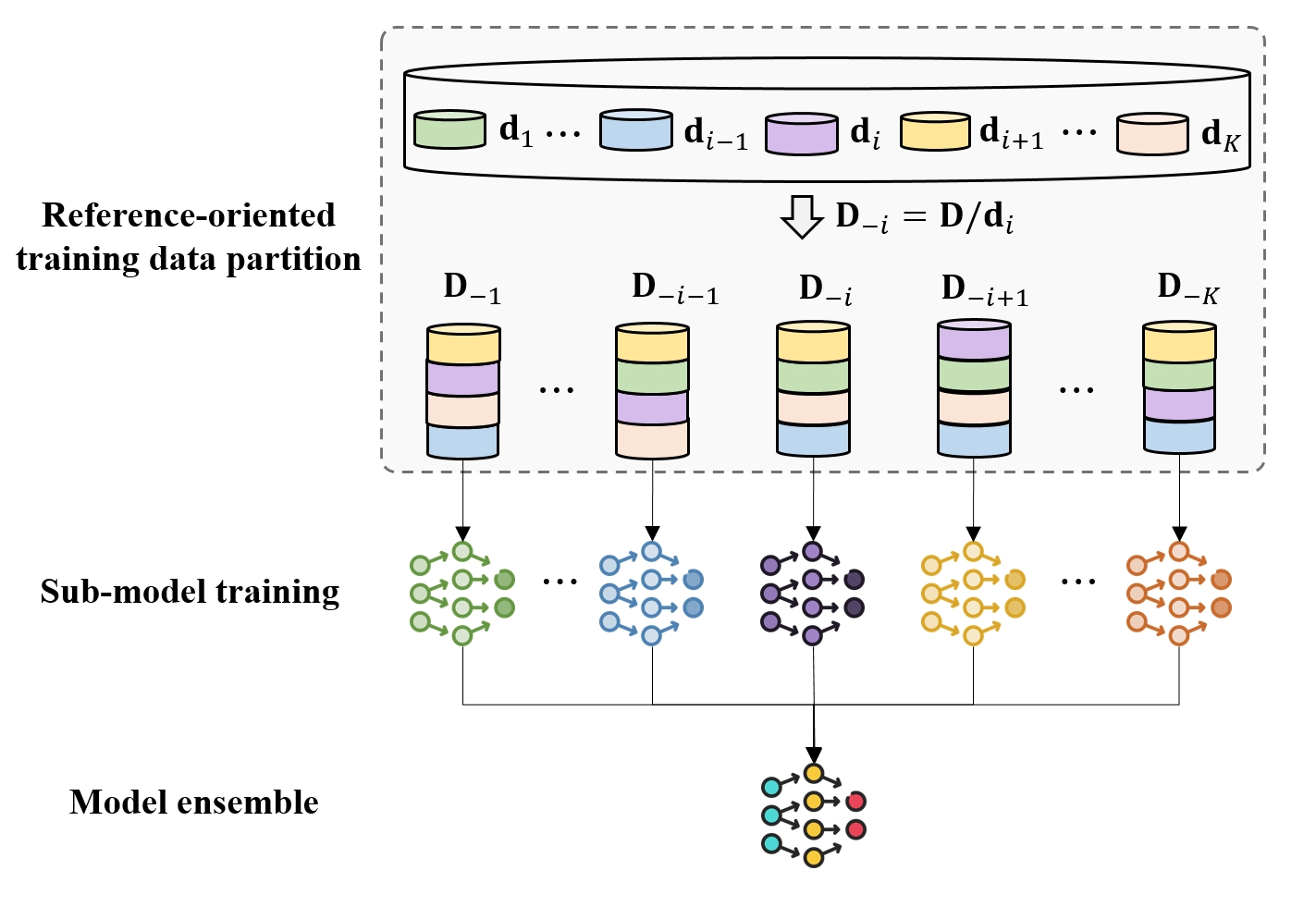}
\caption{Overview of Reference-Oriented Ensemble Learning (ROEL).} \label{fig:roel}
\end{center}
\end{figure}

Given these definitions, ROEL seeks to ensure that any of its generated sub-models can mutually serve as a retrained-alike model for the others, providing suitable reference models for distillation-based unlearning. In particular, ROEL begins by randomly partitioning the training dataset $\mathbf{D}$ into $K$ ($K\geq3$) non-overlapping parts of equal size, denoted as $\mathbf D = {\mathbf d_1} \cup {\mathbf d_2} \cup \cdots \cup {\mathbf d_K}$, where ${\mathbf d_i} \cap {\mathbf d_j} = \emptyset$, for $\forall i,j\in\{1,2,\cdots,K\},i \neq j$. Unlike the existing unlearning method SISA~\citep{bourtoule21} directly using each data part to train a sub-model, ROEL excludes one part ${\mathbf d_i}$ at a time and combines the remaining $K-1$ parts into a subset $\mathbf D_{-i}=\mathbf D / \mathbf d_i$ to train a sub-model. In this manner, ROEL generates $K$ data subsets $\mathbf D_{-i}, i\in\{1,2,\cdots,K\}$ and subsequently trains $K$ sub-models $M_i=A(\mathbf D_{-i})$ using the same algorithm $A$. This approach ensures that each sub-model is trained on most of the training data, thereby enhancing overall model performance. Additionally, it guarantees that any two sub-models are at least $1$-alike, meaning the amount of shared training data between them is larger than the unique data each excludes. Finally, the output of the ensemble predictive model is obtained by averaging the predictions of the $K$ sub-models: $M = \frac{1}{K} \sum_{i=1}^{K} M_i$. The overview of ROEL is illustrated in Figure~\ref{fig:roel}.

\begin{proposition}
\label{proposition1}
    Given the sub-models generated by ROEL and unlearning data $\mathbf D^u \subset \mathbf d_i$, sub-model $M_i$ is a retrained-alike model of sub-model $M_j, \forall j \neq i$.
\end{proposition}
Proof. See Appendix A.1.

In the following we elaborate on the novelty and advantages of our ROEL method in comparison with the previous ensemble methods designed for unlearning. The previous method~\citep{bourtoule21} aims to facilitate partial retraining by constructing numerous small sub-models, where each data sample is only associated with one sub-model. To facilitate efficient partial retraining, this method requires each sub-model to be trained on a small subset of the data, which often conflicts with the goal of producing accurate sub-models by using a sufficient amount of training samples. In contrast, our method is specifically designed to provide suitable reference models for distillation-based unlearning by constructing retrained-like models during the initial model construction stage without additional effort. By ensuring that each sub-model is adequately trained using most of the overall training data, our method achieves an accurate predictive model. At the same time, these sub-models through careful design can mutually serve as retrained-like reference models to effectively support unlearning.

\subsection{Iterative Information Distillation}
\label{sec:IID}
\begin{figure}[t]
\begin{center}
\includegraphics[width=0.97\linewidth]{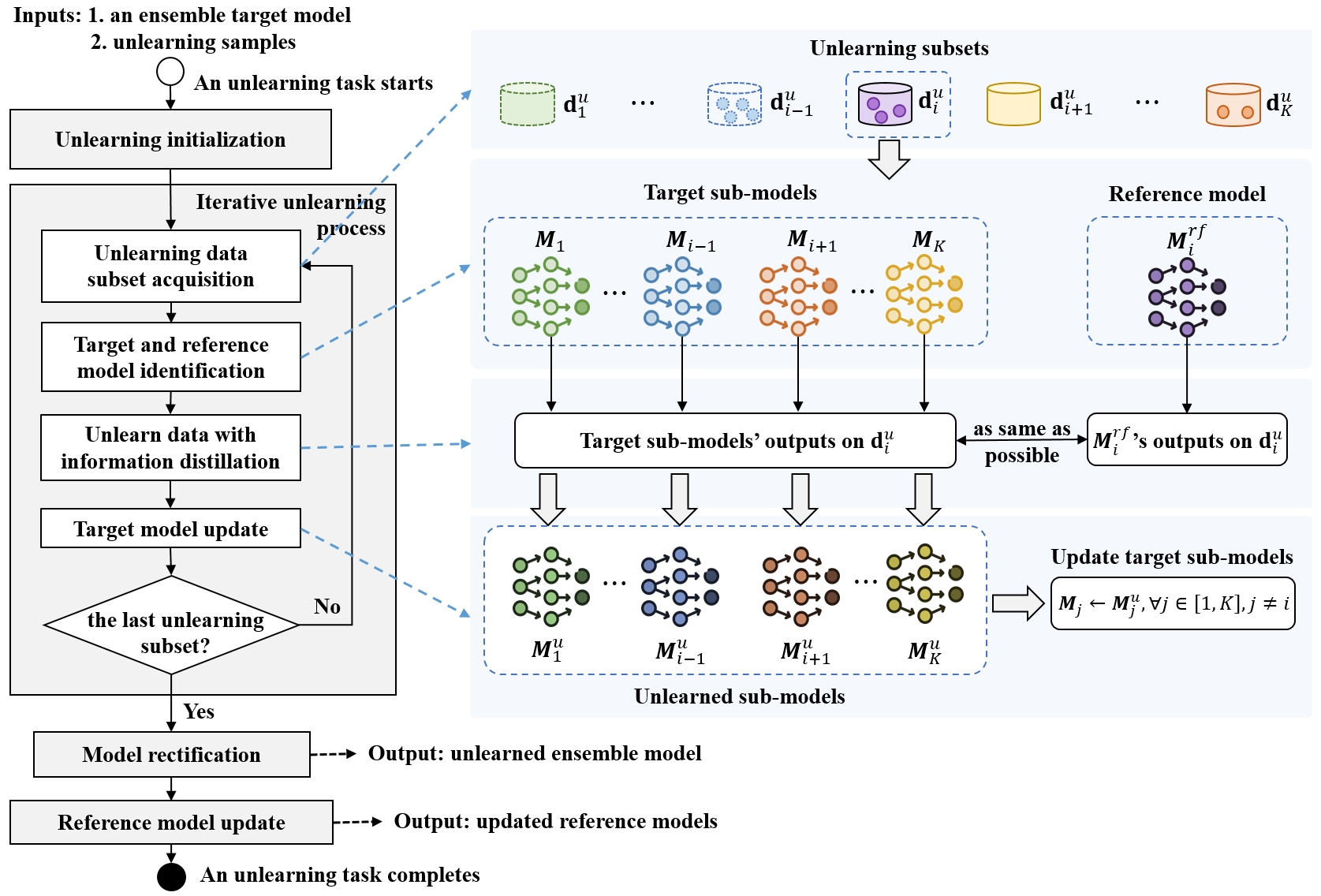}
\caption{The flow diagram of iTerative Information Distillation (TID).} \label{fig:iid_diagram}
\end{center}
\end{figure}

We develop a new distillation-based method named i\textit{T}erative \textit{I}nformation \textit{D}istillation (TID) to handle unlearning requests of data subjects in the unlearning response stage. Benefited from our ROEL design, we are able to obtain retrained-alike reference models for our distillation process to guarantee the consistency of unlearning results without incurring additional computational costs. When a request is made to unlearn data $\mathbf{D}^u$, which may include multiple samples distributed across different data parts generated by ROEL, TID begins by grouping the unlearning samples within each part into separate unlearning subsets. The method then iteratively distills the information of each unlearning subset from the corresponding sub-models (\emph{i.e.}, target sub-models). Following this, TID rectifies the unlearned model through additional distillation steps to enhance its predictive performance and then updates the reference models. Figure~\ref{fig:iid_diagram} presents the flow diagram of TID.

\subsubsection{Unlearning Initialization.} To initialize the unlearning process, we first construct unlearning data subsets and prepare the reference model for each subset. The unlearning data $\mathbf{D}^u$ is partitioned into $K$ unlearning subsets $\mathbf{d}_i^u \subset \mathbf{d}_i$ based on the samples’ membership in the $K$ data parts generated by ROEL. Specifically, all unlearning samples that belong to data part $\mathbf{d}_i$ are grouped into unlearning subset $\mathbf{d}_i^u$. We denote $\mathbf{D}^u = \mathbf{d}_1^u \cup \mathbf{d}_2^u \cup \cdots \cup \mathbf{d}_K^u$, where $\mathbf{d}_i^u = \emptyset$ if no unlearning data is present in data part $\mathbf{d}_i$.

It is important to note that for any non-empty $\mathbf{d}_i^u$, all sub-models $\{M_j | j \neq i\}$ generated by ROEL have been trained on $\mathbf{d}_i^u$ and therefore must undergo unlearning. According to Proposition~\ref{proposition1}, sub-model $M_i$ can serve as a retrained-alike model of arbitrary sub-model $M_j, \forall j \neq i$ for the unlearning data $\mathbf{d}_i^u$. Consequently, $M_i$ is identified as the reference model for unlearning $\mathbf{d}_i^u$ from sub-model $M_j, \forall j \neq i$ through distillation-based unlearning. Then sub-model $M_i$ is duplicated to create the reference model $M^{\textit{rf}}_i \leftarrow M_i$ for unlearning $\mathbf{d}_i^u$.

\subsubsection{Iterative Unlearning Process.} The unlearning process is conducted iteratively by distilling the information of each unlearning subset. Specifically, each iteration involves several steps. First, we acquire a non-empty unlearning subset $\mathbf{d}_i^u$ and identify the corresponding reference model $M^{\textit{rf}}_i$ along with the target sub-models for distillation, denoted as $\mathbb M^{tg}_i=\{M_1,\cdots,M_{i-1},M_{i+1},\cdots,M_K\}$. We then unlearn $\mathbf{d}_i^u$ by distilling its information from each target sub-model $M_j \in \mathbb{M}^{tg}_i$ under the supervision of $M^{\textit{rf}}_i$. The goal of distillation is to align the predictions of the target sub-models on the unlearning data with those of the reference model. Notably, current RTBF regulations do not prohibit the use of unlearning data $\mathbf{D}^u$ when processing unlearning requests, making it permissible for TID to leverage this unlearning data for distillation-based information erasure~\citep{golatkar21,kurmanji23}. This objective is formalized by the following optimization problem:
\begin{equation}
\label{eq:unlearning}
    M^u_j = \min_{M_j} \textsc{diff}(M^{\textit{rf}}_i(\mathbf X_{\mathbf d^u_i}),M_j(\mathbf X_{\mathbf d^u_i})), \quad \forall M_j \in \mathbb M^{tg}_i,
\end{equation}
where $\mathbf X_{\mathbf d^u_i}$ represents the feature set of samples in $\mathbf{d}_i^u$; $M^{\textit{rf}}_i(\mathbf X_{\mathbf d^u_i})$ and $M_j(\mathbf X_{\mathbf d^u_i})$ denote the output posterior probability distributions for the predictive task labels of $M^{\textit{rf}}_i$ and $M_j$, respectively; and $\textsc{diff}(\cdot)$ is a general function that measures the difference between the outputs. By optimizing E.q.~\ref{eq:unlearning}, an unlearned model is obtained and used to update the original target sub-model, \emph{i.e.}, $M_{j} \leftarrow M^{u}_{j}$.

\subsubsection{Model Rectification.}
\label{sec:rectify}
Some unlearning data samples may retain unique information on the target model, while others may share common information with remaining data. During the distillation process, the shared information that can be learned from the remaining data might be inadvertently erased, potentially diminishing model accuracy. To preserve accuracy after unlearning, we rectify the model using the remaining data $\mathbf{D}^r = \mathbf{D} / \mathbf{D}^u$. Specifically, we distill the unlearned sub-models under the supervision of the actual labels of the remaining data, allowing the models to relearn the erased information of the remaining data:
\begin{equation}
\label{eq:rectification}
    M^{*}_j = \min_{M_j} \textsc{diff}(\mathbf Y_{\mathbf D^r_{-j}},M_j(\mathbf X_{\mathbf D^r_{-j}})), \quad \forall j \in [1,K],
\end{equation}
where $\mathbf D^r_{-j}=\mathbf D_{-j} / \mathbf D^u$ represents the remaining data for sub-model $M_j$, and $\mathbf Y_{\mathbf D^r_{-j}}$ and $\mathbf X_{\mathbf D^r_{-j}}$ denote the labels and features of the remaining samples, respectively. With the rectification process, the unlearned target sub-models are further updated to rectified versions, \emph{i.e.}, $M_{j} \leftarrow M^{*}_{j}$.


\subsubsection{Reference model update.}
The reference models are duplicates of the original sub-models that involve the unlearning data. At the end of the unlearning process, we update the reference models by replacing them with the corresponding unlearned sub-models:
\begin{equation}
\label{eq:reference_update}
    M^{\textit{rf}}_i = M_i, \quad \forall i\in[1,K]\ \& \ \mathbf d^u_i\neq \emptyset.
\end{equation}

\begin{proposition}
\label{proposition2}
    Given the unlearned sub-models $\{M_{j}|j\in[1,K]\}$, updated reference models $\{M^{rf}_{i}|i\in[1,K]\}$, and new coming unlearning data $\mathbf D^{u}{'} \subset \mathbf d_i$, $M^{rf}_{i}$ can still be considered a retrained-alike model of sub-model $M_j, \forall j \neq i$, if and only if: 
    \begin{equation}
    \label{eq:prop2_condition}
    \frac{|\mathbf D^{r}_{-i} \bigcap \mathbf D^{r}_{-j}|}{\max(|\mathbf D^{r}_{-i}|-|\mathbf D^{r}_{-i} \bigcap \mathbf D^{r}_{-j}|,|\mathbf D^{r}_{-j}|-|\mathbf D^{r}_{-i} \bigcap \mathbf D^{r}_{-j}|)}\geq 1.
    \end{equation}
    where $\mathbf D^r_{-i}=\mathbf D_{-i} / \mathbf D^u$ and $\mathbf D^r_{-j}=\mathbf D_{-j} / \mathbf D^u$ denote the remaining data of sub-model $M^{rt}_i$ and $M_j$ after prior unlearning, $|\mathbf D^{r}_{-i} \bigcap \mathbf D^{r}_{-j}|$ stands for the number of common training data samples between $M^{rt}_i$ and $M_j$, $|\mathbf D^{r}_{-i}|-|\mathbf D^{r}_{-i} \bigcap \mathbf D^{r}_{-j}|$ and $|\mathbf D^{r}_{-j}|-|\mathbf D^{r}_{-i} \bigcap \mathbf D^{r}_{-j}|$ represent the number of unique training data samples for $M^{rt}_i$ and $M_j$, respectively.
\end{proposition}
Proof. See Appendix A.2.

Proposition~\ref{proposition2} outlines the conditions under which updated reference models can continue to function like retrained models for supporting future unlearning. When these conditions no longer apply --- typically due to the erasure of a significant volume of data --- we recommend introducing new data samples and retraining the model from scratch using ROEL. Potential strategies for adjusting the model with the newly added data are further discussed in the Conclusion and Discussion section.

\subsubsection{Paralleled unlearning and rectification.} While distillation techniques significantly reduce computational costs compared to retraining from scratch, our method can be further accelerated through parallel computing. Specifically, the information distillation and target sub-model updates in the iterative unlearning process can be parallelized. Additionally, both the model rectification and reference model update processes can also be executed in parallel.

\subsection{The property analysis on ETID}

Below, we summarize the advantageous properties of ETID and how ETID adequately addresses the machine unlearning issue. ETID is the first holistic machine learning-to-unlearning framework that systematically addresses the unlearning issue at both the model construction and unlearning request response stages. It leverages the advantages of ensemble and distillation learning to overcome limitations in previous machine unlearning approaches. Specifically, in the first stage, ETID introduces the novel ROEL method, which not only constructs a high-performance predictive model but also prepares retrained-alike reference models for subsequent distillation-based unlearning without requiring extra effort, effectively resolving the lack of suitable reference models in earlier distillation-based methods. In the second stage, ETID proposes a new distillation-based approach, TID, specifically designed for the predictive model established by ROEL. TID utilizes the retrained-alike reference models to ensure unlearning \textit{consistency} during information distillation. Unlike partial retraining techniques, distillation-based unlearning allows sub-models to be trained on sufficient data, ensuring the predictive model's accuracy is maintained. Additionally, model rectification is employed to improve the \textit{accuracy} of the unlearned model. TID maintains \textit{efficiency} by fine-tuning the sub-models instead of retraining them from scratch and by utilizing parallel computing techniques.


Following Definition~\ref{def:Unlearning}, ETID is verifiable if its unlearned model is distinguishable from the target model. As ETID generates an ensemble model comprising multiple sub-models, it is verifiable as long as any of its unlearned sub-models is distinguishable from the corresponding target sub-model. Proposition~\ref{proposition3} demonstrates the \textit{verifiability} of ETID.

\begin{proposition}
\label{proposition3}
    Given sub-model $M_j$ and its unlearned sub-model $M^u_j$ generated by ETID, $M^u_j$ is distinguishable from $M_j$, for $\forall j \in [1,K]$.
\end{proposition}
Proof. See Appendix A.3.

\section{Experiments}
In this section, we experimentally validate the superiority of our ETID framework with two datasets. Below, we present our experimental setups and results.

\subsection{Experimental Setups}
\label{sec:exp_setups}

\begin{table*}[t]\footnotesize
\centering
\caption{Summary statistics of evaluation datasets.}
\def\arraystretch{1.3}\begin{tabular}{ccccc}
\toprule
 \textbf{Datasets}    & \textbf{Predictive models}  & \textbf{\# samples}    & \textbf{Labels}  &\textbf{Features}\\  \midrule
 Purchase   & Consumer Profiling &197,324   & 100 consumer classes &  purchase records on 600 products \\  
 CIFAR100 & Image Classification & 60,000   & 100 image classes & 32×32 color images\\  \bottomrule
\end{tabular}%
\label{tab:dataset}
\end{table*}%

In reality, businesses may collect consumers' historical shopping transaction data or images to train predictive models for consumer profiling or image classifications. In this work, we consider unlearning requests for erasing data from consumer profiling and image classification models, respectively. Following prior works~\citep{bourtoule21}, we apply the consumer shopping transactions dataset Purchase to build the profiling model with a 4-layer fully-connected neural network; we employ ResNet18~\citep{he16} to train the image classification model on the image dataset CIFAR100. Table \ref{tab:dataset} describes the statistics of the datasets. In particular, we randomly split the Purchase data set, 80\% for training and 20\% for testing; as for CIFAR100, we follow the commonly used setting to use 50,000 images for training and the remaining 10,000 images for testing.

\begin{table*}[t]\footnotesize
\centering
\caption{Unlearning methods for comparisons.}
\def\arraystretch{1.3}\begin{tabular}{ccl}
\toprule
&\textbf{Category}   & \multicolumn{1}{c}{\textbf{Method/Framework}}  \\  \midrule
\multirow{6}{*}{Benchmarks} & Theory-based   & Fisher~\citep{golatkar20a} \\  \cline{2-3}
 & Distillation-based & \begin{tabular}[l]{@{}l}Relabel~\citep{graves21}\\Forsaken~\citep{ma22}\\Bad-T~\citep{chundawat23}\\SCRUB~\citep{kurmanji23}\end{tabular}\\ \cline{2-3}
 & Ensemble-based & SISA~\citep{bourtoule21} \\ \midrule
 Our framework & Ensemble \& Distillation & \multicolumn{1}{c}{ETID} \\ \bottomrule
\end{tabular}%
\label{tab:benchmark}
\end{table*}%

Table~\ref{tab:benchmark} lists all the unlearning benchmarks to be compared with. As mentioned, while some existing methods design post-modification unlearning algorithms for a general single predictive model, others devise predictive model construction approaches to facilitate their responses to unlearning requests; our method provides a holistic framework from modeling to unlearning. In the experiments, we follow various designs of unlearning methods to train suitable target models for them. Specifically, we train a single target model (\emph{i.e.}, Target-Single) for benchmarks including Fisher, Relabel, Forsaken, Bad-T, and SCRUB; SISA and ETID propose their ensemble target model construction methods. Thus, we follow the SISA and ETID train two ensemble target models, respectively, which are denoted as Target-SISA and Target-ETID. Retrain-Single and Retrain-ETID denote the na\"ive retrained models of Target-Single and Target-ETID, respectively, used as references for efficiency comparisons. By adopting a partial retraining technique for unlearning, SISA itself is the na\"ive retrained model of Target-SISA.

Following the common practice, we utilized Kullback–Leibler divergence as the difference measure $\textsc{diff}(\cdot)$ in distillation objective functions E.q.~\ref{eq:unlearning} and E.q.~\ref{eq:rectification}~\citep{ma22}. We randomly selected 1.0\% of the training data as the unlearning samples~\citep{bourtoule21} and set the number of sub-models (\emph{i.e.,} $K$) for the ensemble methods as 5 in the main experiments~\citep{breiman96}. We also varied these parameters for parameter sensitivity analysis. All experiments were conducted on an Ubuntu 18.04 server with an Intel(R) Xeon(R) silver 4210 CPU, 256 GB RAM, and a Tesla V100S GPU with 32 GB memory. All methods were implemented with Python 3.8.0 and Pytorch 1.7.0. We repeated each experiment five times and reported the mean and standard deviation results.

\subsection{Experimental Results}
\label{sec:results}

In this section, we first present the prediction performance of target models created by different model construction methods in Section~\ref{sec:pred_pfm_cmp}. More importantly, we examine the unlearning performance of various unlearning methods under the default experimental settings across unlearning desiderata, including \textit{consistency}, \textit{accuracy}, \textit{efficiency}, and \textit{verifiability},  from Section~\ref{sec:unlearning_con} to~\ref{sec:unlearning_ver}. Lastly, we report additional results related to the parameter sensitivity analysis of our proposed framework in Section~\ref{sec:para_sensitivity}.

\subsubsection{Prediction performance comparison of different target models.}
\label{sec:pred_pfm_cmp}In this part, we design experiments to demonstrate the superior prediction performance of the target model constructed using our ROEL method. In particular, we compare the prediction performance of target models established by various model construction methods (\emph{e.g.}, Target-Single, Target-SISA and Target-ETID) on both the consumer profiling and the image classification tasks. It is worth noting that for each task, we apply the same machine learning algorithm to all target models (as described in Section~\ref{sec:exp_setups}), since our focus is on comparing the model construction methods. We use accuracy as the metric to evaluate the prediction performance of the models across various datasets, including remaining data, testing data, and unlearning data. Specifically, accuracy measures the ratio of correct predictions made by the model, defined as follows:
\begin{equation}
\label{eq:acc}
    Acc(\mathbf{\hat{X}}) =  \frac{\sum_{i=1}^{|\mathbf{\hat{X}}|} \mathbb{I}(\hat{y}_i=y_i)}{|\mathbf{\hat{X}}|}
\end{equation}
where $\mathbf{\hat{X}} \in \{\mathbf X^{r}, \mathbf X^{t}, \mathbf X^{u}\}$, $\mathbb{I}(\cdot)$ is a indicator function, $\hat{y}_i$ is the model's predicted task label with input features $x_i$, and $y_i$ is the actual task label. A higher value of $Acc(\cdot)$ indicates better prediction performance, implying that the target model can provide superior predictive services.

\begin{table*}[t]\footnotesize
\centering
\caption{Accuracy evaluation results of target models.}
\def\arraystretch{1.3}\begin{tabular}{c|ccc|ccc}
\toprule
  & \multicolumn{3}{c|}{\textbf{Purchase}}  & \multicolumn{3}{c}{\textbf{CIFAR100}}  \\  \midrule
 \textbf{Methods}   & $Acc(\mathbf X^{r})$ & $Acc(\mathbf X^{t})$  & $Acc(\mathbf X^u)$  & $Acc(\mathbf X^{r})$ & $Acc(\mathbf X^{t})$  & $Acc(\mathbf X^u)$  \\  \midrule
 Target-Single & \underline{$0.986 \pm 0.003$} & \underline{$0.907 \pm 0.007$}  & \underline{$0.981 \pm 0.004$}  & \pmb{$1.000 \pm 0.000$} & \underline{$0.766 \pm 0.005$} & \pmb{$1.000 \pm 0.000$}  \\  
 Target-SISA & $0.937 \pm 0.003$ & $0.897 \pm 0.004$    & $0.935 \pm 0.008$  & \underline{$0.648 \pm 0.001$} & $0.523 \pm 0.003$  & \underline{$0.646 \pm 0.002$} \\
 Target-ETID & \pmb{$0.999 \pm 0.001$} & \pmb{$0.947 \pm 0.001$}  & \pmb{$0.999 \pm 0.001$}  & \pmb{$1.000 \pm 0.000$} & \pmb{$0.776 \pm 0.004$} & \pmb{$1.000 \pm 0.000$}  \\ 
 \bottomrule
 \multicolumn{7}{l}{\makecell[l]{\textit{Notes.} The best and the second best results are highlighted in \textbf{bold} and \underline{underlined}, respectively.}}
\end{tabular}%
\label{tab:target_accuracy}
\end{table*}%

We list the accuracy results of target models established with different model construction methods in Table~\ref{tab:target_accuracy}. From the results, we can see that Target-ETID using our proposed ensemble learning method ROEL gains much higher accuracy than Target-SISA using the ensemble learning presented in~\cite{bourtoule21}. Moreover, the accuracy of Target-SISA is even lower than that of Target-Single. In detail, for the Purchase dataset, Target-ETID's accuracy with $\mathbf{X}^r$, $\mathbf{X}^t$, and $\mathbf{X}^u$ is 1.32\%, 4.41\%, and 1.83\% higher than that of Target-Single; while they are 6.62\%, 5.57\%, and 6.84\% higher than those of Target-SISA. For the CIFAR100 dataset, Target-ETID's accuracy results again outperform those of other methods. The accuracy results of Target-SISA only achieve $0.523$ on testing data and $0.646$ on unlearning data, which is considerably lower than the accuracy scores achieved by Target-Single. These observations strongly support the conclusion that ROEL, as implemented in Target-ETID, provides a significant advantage in training sub-models with adequate sample sizes, leading to overall higher model accuracy compared to both Target-SISA and Target-Single.

\subsubsection{Unlearning consistency evaluation.} \label{sec:unlearning_con}
This experiment aims to evaluate the consistency of the unlearned model produced by our proposed framework compared to benchmark methods. Specifically, consistency is often measured by the distance in predictions between the unlearned model $M^u$ and the na\"ive retrained model $M^{rt}$. Following the previous work \citep{golatkar20a,chundawat23}, we use L2-distance as the metric of consistency, formally defined as:
\begin{equation}
    Con(\mathbf{\hat{X}}) = \sum^{|\mathbf{\hat{X}}|}_{i=1} ||M^{rt}(\mathbf x_i) - M^u(\mathbf x_i)||_{2}
\end{equation}
We perform na\"ive retraining to obtain $M^{rt}$ and then compare it with $M^u$ derived from an unlearning method to compute $Con(\cdot)$. A smaller value of the metric $Con(\cdot)$ indicates higher consistency.

\begin{table*}[t]\footnotesize
\centering
\caption{Consistency evaluation results}
\def\arraystretch{1.3}\begin{tabular}{c|ccc|ccc}
\toprule
  & \multicolumn{3}{c|}{\textbf{Purchase}}  & \multicolumn{3}{c}{\textbf{CIFAR100}}  \\  \midrule
 \textbf{Methods}   & $Con\mathbf X^r)$   & $Con(\mathbf X^t)$   & $Con(\mathbf X^u)$  & $Con(\mathbf X^r)$   & $Con(\mathbf X^t)$  & $Con(\mathbf X^u)$ \\  \midrule
  Fisher & $0.189 \pm 0.021$    & $0.257 \pm 0.027$  & $0.247 \pm 0.034$ & \underline{$0.009 \pm 0.000$} & \underline{$0.213 \pm 0.000$} & \underline{$0.370 \pm 0.002$}  \\  
  Relabel & $0.060 \pm 0.003$    & $0.162 \pm 0.005$  & $0.392 \pm 0.014$ & $0.018 \pm 0.000$ & $0.215 \pm 0.001$ & $1.280 \pm 0.007$  \\  
  Forsaken & $0.210 \pm 0.006$    & $0.199 \pm 0.006$  & $0.328 \pm 0.013$ & $0.027 \pm 0.001$ & $0.235 \pm 0.002$ & $0.468 \pm 0.010$  \\  
  Bad-T & $0.067 \pm 0.009$    & $0.163 \pm 0.007$  & $0.482 \pm 0.011$ & $0.019 \pm 0.003$ & $0.259 \pm 0.005$ & $0.796 \pm 0.005$  \\  
  SCRUB & \underline{$0.057 \pm 0.006$}    & \underline{$0.160 \pm 0.005$}  & \underline{$0.198 \pm 0.014$} & $0.012 \pm 0.001$ & $0.221 \pm 0.002$ & $0.394 \pm 0.017$  \\  
  ETID & \pmb{$0.045 \pm 0.003$}    & \pmb{$0.087 \pm 0.003$}  & \pmb{$0.141 \pm 0.006$} & \pmb{$0.005 \pm 0.000$} & \pmb{$0.130 \pm 0.003$} & \pmb{$0.204 \pm 0.009$}  \\  \bottomrule
  \multicolumn{7}{l}{\makecell[l]{\textit{Notes.} SISA is not listed in the table as its $Con(\cdot)$ is always zero by retraining target sub-models. The best\\and the second-best results are highlighted in \textbf{bold} and \underline{underlined}, respectively.}}
\end{tabular}%
\label{tab:consistency}
\end{table*}%

Table~\ref{tab:consistency} reports the consistency evaluation results of unlearning methods over remaining data $\mathbf X^{r}$, testing data $\mathbf X^{t}$ and unlearning data $\mathbf X^{u}$. The results show that our method ETID derives much lower consistency metric values than all the other non-retrained methods on both Purchase and CIFAR100. For instance, as the best non-retrained benchmark method on Purchase, SCRUB's consistency metric values over training, testing, and unlearning data are 26.7\%, 83.9\% and 40.4\% larger than those of ETID, which demonstrates the superior consistency of ETID. Additionally, on the CIFAR100 dataset, ETID also achieves the best consistency results, with the metric values being 0.005, 0.130, and 0.204 over different data components. In comparison, SCRUB's consistency metric values are substantially higher, with the values over testing and unlearning data being 70.0\% and 93.1\% larger than those of ETID, respectively. The results also reveal the evident inconsistency of Relabel and Bad-T over unlearning data, as these methods adopt unsuitable reference models. For instance, Bad-T exhibits particularly high consistency metric values on both datasets, which indicates poor consistency performance. Besides, we notice that the theory-based method Fisher attains a noticeably high actual consistency on CIFAR100 but struggles with lower consistency on Purchase. In contrast, Forsaken shows a moderate performance across both datasets but does not match the consistency levels achieved by ETID.

In all, this evaluation demonstrates that for distillation-based unlearning methods, finding an appropriate reference model is crucial to ensure the consistency of the unlearned model. It also validates that ETID can attain the highest consistency in addressing unlearning requests among all non-retrained methods, ensuring lower $Con(\cdot)$ values across various data types and datasets.

\subsubsection{Unlearning accuracy evaluation.} \label{sec:unlearning_acc}
In this experiment, we aim to examine whether the unlearned model produced by our proposed framework, ETID, can provide better prediction performance compared to benchmarks. Specifically, we first use ETID and benchmarks to obtain the corresponding unlearned models and then evaluate their prediction performance on the consumer profiling and image classification tasks. We also use accuracy defined in E.q.~\ref{eq:acc} as the metric to assess the prediction performance of unlearned models.

Table~\ref{tab:accuracy} outlines the accuracy results of unlearned models using various unlearning methods. We observe that our framework ETID significantly outperforms all the benchmark methods by achieving the highest accuracy results over remaining data $\mathbf{X}^r$, testing data $\mathbf{X}^t$, and unlearning data $\mathbf{X}^u$ on both Purchase and CIFAR100 datasets. In particular, on the Purchase dataset, ETID achieves perfect accuracy result on the remaining data with 1.000. For the testing data, ETID's accuracy is 4.04\% higher than that of SCRUB (the best performing benchmark method), and ETID surpasses SCRUB by 0.22\% in accuracy over the unlearning data. Moreover, on the CIFAR100 dataset, ETID also leads the accuracy performance over various data components. In detail, on the testing data, the accuracy result of our method is 2.24\% higher than that of the best performing benchmark, Relabel. For the unlearning data, ETID achieves the highest accuracy result with 0.776, which is also significantly higher than those accuracy results of all the benchmark methods.

\begin{table*}[t]\footnotesize
\centering
\caption{Accuracy evaluation results.}
\def\arraystretch{1.3}\begin{tabular}{c|ccc|ccc}
\toprule
  & \multicolumn{3}{c|}{\textbf{Purchase}}  & \multicolumn{3}{c}{\textbf{CIFAR100}}  \\  \midrule
 \textbf{Methods}  & $Acc(\mathbf X^{r})$ & $Acc(\mathbf X^{t})$  & $Acc(\mathbf X^u)$  & $Acc(\mathbf X^{r})$ & $Acc(\mathbf X^{t})$  & $Acc(\mathbf X^u)$  \\  \midrule 
 Fisher & $0.891 \pm 0.006$ & $0.837 \pm 0.006$  & $0.886 \pm 0.009$ & $0.938 \pm 0.008$ & $0.477 \pm 0.001$ & $0.624 \pm 0.003$   \\  
 Relabel & \underline{$0.993 \pm 0.002$} & $0.910 \pm 0.007$   & $0.715 \pm 0.015$  & \pmb{$1.000 \pm 0.000$} & \underline{$0.759 \pm 0.002$} & $0.011 \pm 0.003$  \\  
 Forsaken & $0.901 \pm 0.007$ & $0.901 \pm 0.005$   & $0.795 \pm 0.013$  & $0.711 \pm 0.004$ & $0.721 \pm 0.002$ & $0.657 \pm 0.017$  \\  
 Bad-T & $0.991 \pm 0.007$ & $0.915 \pm 0.006$    & $0.738 \pm 0.016$  & \underline{$0.998 \pm 0.001$} & $0.735 \pm 0.004$ & $0.362 \pm 0.033$   \\  
 SCRUB & $0.968 \pm 0.009$ & \underline{$0.915 \pm 0.003$}    & \underline{$0.910 \pm 0.003$}  & \pmb{$1.000 \pm 0.000$} & $0.757 \pm 0.004$ & \underline{$0.727 \pm 0.026$}  \\  
 SISA & $0.907 \pm 0.004$ & $0.896 \pm 0.004$    & $0.895 \pm 0.010$  & $0.657 \pm 0.006$ & $0.533 \pm 0.001$  & $0.560 \pm 0.021$   \\  
 ETID & \pmb{$1.000 \pm 0.000$} & \pmb{$0.952 \pm 0.001$}     & \pmb{$0.912 \pm 0.001$}  & \pmb{$1.000 \pm 0.000$} & \pmb{$0.776 \pm 0.003$} & \pmb{$0.776 \pm 0.015$}  \\   
 \bottomrule
 \multicolumn{7}{l}{\makecell[l]{\textit{Notes.} The best and the second-best results are highlighted in \textbf{bold} and \underline{underlined}, respectively.}}
\end{tabular}%
\label{tab:accuracy}
\end{table*}%

Additionally, we observe that SISA presents poor accuracies, especially in the complex learning task on CIFAR100. The reason is that SISA splits the training data into distinct subsets with limited data to facilitate partial retraining, which easily causes insufficient training of its sub-models. For instance, the accuracy results of unlearned model produced by SISA are 0.657 and 0.533 on CIFAR100, which are notably lower compared to those of ETID. Besides, Relabel shows a noticeable drop in accuracy results on unlearning data on Purchase, which highlights its defect in maintaining high accuracy after responding to unlearning requests. Bad-T also demonstrates substantial accuracy losses on unlearning data, particularly on CIFAR100. These results emphasize the importance of selecting an appropriate reference model for distillation-based unlearning methods. It is crucial not only for maintaining the consistency of the unlearned model but also has a significant impact on the accuracy performance after responding to unlearning requests.

Overall, our proposed framework ETID demonstrates superior accuracy performance compared to other unlearning methods, and the superiority is even more significant in the more complex predictive task on CIFAR100, as evidenced by the larger gaps in accuracy metrics. These findings suggest that business companies can effectively offer accurate predictive services by using ETID to address unlearning requests. Such accurate predictive services can, in turn, help companies increase profitability by better meeting customer needs and optimizing decision-making processes.

\subsubsection{Unlearning efficiency evaluation.}
\label{sec:unlearning_eff}
This experiment evaluates the efficiency of the unlearning process within the ETID framework compared to benchmark methods. Specifically, we maintain the same running environment and record the time expense for each method's unlearning process. We then compare their efficiency based on these time costs. Generally, the time cost of an unlearning method is expected to be significantly lower than that of retraining the predictive model from scratch. The lower the time cost, the higher the efficiency.

Table \ref{tab:efficiency} displays the running time of various unlearning methods. In general, distillation-based methods are much faster than theory-based methods and SISA. ETID further accelerates the unlearning procedure and achieves optimal efficiency by employing distillation techniques with parallel computing. Notably, ETID reduces the unlearning time by 80.2\% and 88.4\% compared to the single naïve retrained model (Retrain-Single) on Purchase and CIFAR100, respectively. Specifically, on the Purchase dataset, ETID achieves a running time of $43.51$ seconds when using serial computation and $8.63$ seconds with parallel computing. This is a significant reduction compared to Retrain-Single, which takes $43.59$ seconds. In contrast, the theory-based method Fisher takes a much longer time to complete the unlearning process, demonstrating dramatically higher computational overhead. Similarly, Bad-T incurs higher time costs than Retrain-Single at $111.06$ seconds, making it less practical for efficient unlearning. On the CIFAR100 dataset, ETID's efficiency is even more pronounced. It achieves a running time of $958.23$ seconds with parallel computing. In comparison, Retrain-Single requires $8295.03$ seconds. The distillation-based methods such as SCRUB also show competitive performance with $1137.47$ seconds but still fall short of ETID's efficiency. Notably, Fisher again shows significant inefficiency. As for SISA, it consumes $8745.14$ seconds to erase the unlearning data, highlighting its impracticality for time-efficient unlearning.

\begin{table*}[t]\footnotesize
\centering
\caption{Efficiency evaluation results.}
\def\arraystretch{1.3}\begin{tabular}{c|c|c}
\toprule
   & \textbf{Purchase}  & \textbf{CIFAR100}  \\  \midrule
  \textbf{Methods}   & Running time (s)  & Running time (s)  \\  \midrule
  Retrain-Single & $43.59 \pm 0.69$  & $ 8295.03 \pm 3.65$  \\  
  Retrain-ETID & $180.92 \pm 1.92$   & $47764.46 \pm 49.42$  \\ 
  Fisher & $39888.06 \pm 1683.05$  & $65406.89 \pm 1897.23$   \\  
  Relabel & $33.92 \pm 1.24$   & $8402.01 \pm 4.81$   \\  
  Forsaken & $13.28 \pm 0.69$   & $1254.06 \pm 10.72$   \\  
  Bad-T & $111.06 \pm 1.54$    & $1425.58 \pm 9.59$    \\  
  SCRUB & \underline{$9.21 \pm 0.67$}    & \underline{$1137.47 \pm 54.11$}  \\ 
  SISA & $119.32 \pm 3.92$    & $8745.14 \pm 54.31$    \\  
  ETID (serial/parallel) & $43.51 \pm 0.95$ / \pmb{$8.63 \pm 0.24$}    & $4679.37 \pm 35.65$ / \pmb{$958.23 \pm 5.34$}  \\  \bottomrule
  \multicolumn{3}{l}{\makecell[l]{\textit{Notes.} Retrain-Single and Retrain-ETID are the na\"ive retrained model of Target-Single\\and Target-ETID. The ideal unlearning methods should be much faster than them. The\\best and the second-best results are highlighted in \textbf{bold} and \underline{underlined}, respectively.}}
\end{tabular}%
\label{tab:efficiency}
\end{table*}%

These results indicate that some benchmarks, such as Fisher and SISA, consume more time than the naïve retraining method, rendering them impractical for real-world applications where time efficiency is critical. The superiority of ETID in terms of efficiency makes it a highly practical solution for addressing unlearning requests by ensuring minimal computational overhead among the unlearning methods.

\subsubsection{Unlearning verifiability evaluation.} 
\label{sec:unlearning_ver}

In this experiment, 
we follow previous works to examine the verifiability of various unlearning methods by inferring the membership of the unlearning samples~\citep{ma22,xu23}. Specifically, membership inference is performed on both the target and unlearned models to discern unlearning data as members of training data and testing data as non-members. The verifiability of an unlearning method can be confirmed by a significant difference in membership inference performance between the unlearned and target models. We employ the Area Under the Curve (AUC) score to evaluate membership inference performance (denoted as M-AUC) and compute the M-AUC difference $|\Delta|$ between a target model and its unlearned model derived from an unlearning method. A significant $|\Delta|$ signifies the verifiability of the unlearning method. Detailed implementations of membership inference are presented in Section B of Appendix.

Table \ref{tab:verification} summarizes the results of verifiability examinations. We note that all evaluated unlearning methods exhibit verifiability as they all achieve a significant M-AUC difference $|\Delta|$ between their target and unlearned models. Moreover, it is worth noting that an M-AUC nearing 0.5 indicates that unlearning data cannot be distinguished from testing data through membership inference. We notice some of the benchmarks yield a much lower M-AUC than 0.5. For instance, Bad-T shows an M-AUC of $0.195$ and $0.168$ after unlearning on Purchase and CIFAR100 datasets, respectively. This outcome occurs when membership inference identifies nearly all unlearning data as non-members but misclassifies some testing data as members. These findings suggest these methods likely over-unlearn the data by using some stochastic references and also interpret their low prediction accuracy for unlearning data.

\begin{table*}[t]\footnotesize
\centering
\caption{Unlearning verifiability examination results.}
\def\arraystretch{1.3}\begin{tabular}{c|ccc|ccc}
\toprule
  & \multicolumn{3}{c|}{\textbf{Purchase}}  & \multicolumn{3}{c}{\textbf{CIFAR100}}  \\  \midrule
  \textbf{Methods}   & \makecell{M-AUC \\before unlearning} & \makecell{M-AUC \\after unlearning}  & $|\Delta|$  & \makecell{M-AUC \\before unlearning} & \makecell{M-AUC \\after unlearning}  & $|\Delta|$  \\  \midrule
  Fisher & $0.573 \pm 0.010$ & $0.533 \pm 0.007$  & $0.040^{**}$  & $0.787 \pm 0.015$ & $0.528 \pm 0.019$ & $0.259^{***}$   \\  
  Relabel & $0.573 \pm 0.010$ & $0.314 \pm 0.008$   & $0.259^{***}$  & $0.787 \pm 0.015$ & $0.648 \pm 0.029$ & $0.139^{**}$  \\  
  Forsaken & $0.573 \pm 0.010$ & $0.402 \pm 0.003$   & $0.171^{***}$  & $0.787 \pm 0.015$ & $0.536 \pm 0.014$ & $0.251^{***}$  \\  
  Bad-T & $0.573 \pm 0.010$ & $0.195 \pm 0.016$    & $0.378^{***}$  & $0.787 \pm 0.015$ & $0.168 \pm 0.029$ & $0.619^{***}$   \\  
  SCRUB & $0.573 \pm 0.010$ & $0.521 \pm 0.009$    & $0.052^{**}$  & $0.787 \pm 0.015$ & $0.554 \pm 0.016$ & $0.233^{***}$  \\  
  SISA & $0.535 \pm 0.010$ & $0.496 \pm 0.014$    & $0.039^{**}$ & $0.543 \pm 0.012$  & $0.501 \pm 0.009$  & $0.042^{***}$   \\ 
  ETID & $0.596 \pm 0.013$ & $0.519 \pm 0.008$     & $0.077^{***}$ & $0.772 \pm 0.018$  & $0.544 \pm 0.010$ & $0.228^{***}$   \\  \bottomrule
 \multicolumn{2}{l}{\textit{Notes.} ***$p \leq 0.001$; **$p \leq 0.01$.}
 
\end{tabular}%
\label{tab:verification}
\end{table*}%

\subsubsection{Parameter sensitivity analysis.}
\label{sec:para_sensitivity}
We design experiments to investigate the parameter sensitivity of the proposed unlearning framework by altering the default number of sub-models ($K$) and the ratio of the unlearning samples to the whole training set (\emph{i.e.}, unlearning ratio, abbreviated as UR). Concretely, we keep other default settings unchanged, and conduct Experiment 1 to Experiment 4 with ETID by varying values of $K$ among $\{3,5,7,10\}$, and varying the unlearning ratio among $\{0.1\%, 0.5\%, 1.0\%, 5.0\%, 10.0\%\}$, respectively. 

We report the unlearning performance of ETID with different values of $K$ on Purchase and CIFAR100 in Table~\ref{tab:purchase_k} and Table~\ref{tab:cifar_k}, respectively. It is worth noting that as the number of sub-models $K$ increases, each sub-model has a larger training data size, and they share more training samples. From the results, we observe that the consistency metric decreases as $K$ increases on both datasets, which means that a larger $K$ leads to higher consistency. This is reasonable since as $K$ increases, the reference models we established are more similar to the target sub-model's corresponding na\"ive retraining models due to more shared training samples, thereby demonstrating a higher consistency. Besides, we also note that the accuracy of the unlearned model derived from ETID improves as $K$ increases. This can be attributed to two reasons: first, the increased size of subsets leads to more accurate sub-models; second, the increased number of sub-models results in a more powerful ensemble model. Despite the improvements, a larger $K$ also comes with higher computational and storage costs. Therefore, it is important for model providers to select an appropriate $K$ based on their needs and computational resources.

\begin{table*}[t]\footnotesize
\centering
\caption{Parameter sensitivity analysis of the number of sub-models (K) on Purchase dataset.}
\def\arraystretch{1.3}\begin{tabular}{c|ccc|ccc|c|c}
\toprule
 & \multicolumn{3}{c|}{\textbf{Consistency}}  & \multicolumn{3}{c|}{\textbf{Accuracy}} & \textbf{Efficiency} &  \textbf{Verifiability} \\  \midrule
$K$   & $Con(\mathbf X^{r})$  & $Con(\mathbf X^{t})$  & $Con(\mathbf X^{u})$  & $Acc(\mathbf X^{r})$ & $Acc(\mathbf X^{t})$ & $Acc(\mathbf X^{u})$ & seconds &  $|\Delta|$ \\  \midrule
3 & $0.061 \pm 0.001$  & $0.107 \pm 0.001$  &  $0.134 \pm 0.004$ & $1.000 \pm 0.000$ & $0.947 \pm 0.001$ & $0.907 \pm 0.005$ & $7.36 \pm 0.30$ & $0.072^{***}$ \\ 
5 & $0.045 \pm 0.003$  & $0.087 \pm 0.003$  &  $0.141 \pm 0.006$ & $1.000 \pm 0.000$ & $0.952 \pm 0.001$ & $0.912 \pm 0.001$ & $8.63 \pm 0.24$ & $0.077^{***}$ \\ 
7 & $0.039 \pm 0.001$  & $0.079 \pm 0.002$  &  $0.134 \pm 0.007$ & $1.000 \pm 0.000$ & $0.955 \pm 0.001$ & $0.906 \pm 0.006$ & $9.47 \pm 0.35$ & $0.083^{***}$ \\ 
10 & $0.035 \pm 0.002$  & $0.071 \pm 0.003$  &  $0.135 \pm 0.008$ & $1.000 \pm 0.000$ & $0.956 \pm 0.001$ & $0.906 \pm 0.006$ & $9.68 \pm 0.31$ & $0.102^{***}$ \\ \bottomrule
 \multicolumn{3}{l}{\textit{Notes.} ***$p \leq 0.001$; **$p \leq 0.01$.}
\end{tabular}%
\label{tab:purchase_k}
\end{table*}%

\begin{table*}[t]\footnotesize
\centering
\caption{Parameter sensitivity analysis of the number of sub-models (K) on CIFAR100 dataset.}
\def\arraystretch{1.3}\begin{tabular}{c|ccc|ccc|c|c}
\toprule
 & \multicolumn{3}{c|}{\textbf{Consistency}}  & \multicolumn{3}{c|}{\textbf{Accuracy}} & \textbf{Efficiency} & \textbf{Verifiability} \\  \midrule
$K$   & $Con(\mathbf X^{r})$  & $Con(\mathbf X^{t})$  & $Con(\mathbf X^{u})$  & $Acc(\mathbf X^{r})$ & $Acc(\mathbf X^{t})$ & $Acc(\mathbf X^{u})$ & seconds &  $|\Delta|$  \\  \midrule
3 & $0.004 \pm 0.000$  & $0.155 \pm 0.002$  &  $0.218 \pm 0.006$ & $1.000 \pm 0.000$ & $0.759 \pm 0.002$ & $0.761 \pm 0.014$ & $781.92 \pm 1.72$ & $0.197^{***}$ \\ 
5 & $0.004 \pm 0.000$  & $0.130 \pm 0.003$  &  $0.204 \pm 0.009$ & $1.000 \pm 0.000$ & $0.776 \pm 0.003$ & $0.776 \pm 0.015$ & $958.23 \pm 0.49$ &  $0.228^{***}$ \\ 
7 & $0.003 \pm 0.000$  & $0.100 \pm 0.001$  &  $0.183 \pm 0.006$ & $1.000 \pm 0.000$ & $0.792 \pm 0.001$ & $0.787 \pm 0.016$ & $1032.68 \pm 1.91$ &  $0.220^{***}$ \\ 
10 & $0.003 \pm 0.000$  & $0.084 \pm 0.000$  &  $0.177 \pm 0.005$ & $1.000 \pm 0.000$ & $0.798 \pm 0.001$ & $0.809 \pm 0.020$ & $1113.54 \pm 1.54$ &  $0.217^{***}$ \\ \bottomrule
 \multicolumn{3}{l}{\textit{Notes.} ***$p \leq 0.001$; **$p \leq 0.01$.}
\end{tabular}%
\label{tab:cifar_k}
\end{table*}%

\begin{table*}[t]\footnotesize
\centering
\caption{Parameter sensitivity analysis of the unlearning ratio (UR) on Purchase dataset.}
\def\arraystretch{1.3}\begin{tabular}{c|ccc|ccc|c|c}
\toprule
 & \multicolumn{3}{c|}{\textbf{Consistency}}  & \multicolumn{3}{c|}{\textbf{Accuracy}} & \textbf{Efficiency} & \textbf{Verifiability}  \\  \midrule
UR  & $Con(\mathbf X^{r})$  & $Con(\mathbf X^{t})$  & $Con(\mathbf X^{u})$  & $Acc(\mathbf X^{r})$ & $Acc(\mathbf X^{t})$ & $Acc(\mathbf X^{u})$ & seconds & $|\Delta|$ \\  \midrule
0.1\% & $0.043 \pm 0.004$  & $0.086 \pm 0.003$  &  $0.135 \pm 0.022$ & $1.000 \pm 0.000$ & $0.954 \pm 0.001$ & $0.902 \pm 0.016$ & $8.25 \pm 0.20$ & $0.081^{**}$ \\ 
0.5\% & $0.045 \pm 0.004$  & $0.088 \pm 0.004$  &  $0.125 \pm 0.008$ & $1.000 \pm 0.000$ & $0.953 \pm 0.001$ & $0.916 \pm 0.008$ & $8.67 \pm 0.30$ & $0.088^{**}$ \\ 
1.0\% & $0.045 \pm 0.003$  & $0.087 \pm 0.003$  &  $0.141 \pm 0.006$ & $1.000 \pm 0.000$ & $0.952 \pm 0.001$ & $0.912 \pm 0.001$ & $8.63 \pm 0.24$ & $0.077^{***}$ \\ 
5.0\% & $0.051 \pm 0.002$  & $0.091 \pm 0.003$  &  $0.119 \pm 0.002$ & $1.000 \pm 0.000$ & $0.950 \pm 0.001$ & $0.917 \pm 0.002$ & $8.87 \pm 0.29$ & $0.068^{***}$ \\ 
10.0\% & $0.057 \pm 0.002$  & $0.092 \pm 0.002$  &  $0.108 \pm 0.004$ & $1.000 \pm 0.000$ & $0.949 \pm 0.001$ & $0.929 \pm 0.004$ & $9.21 \pm 0.32$ & $0.073^{***}$ \\ \bottomrule
 \multicolumn{3}{l}{\textit{Notes.} ***$p \leq 0.001$; **$p \leq 0.01$.}
\end{tabular}%
\label{tab:purchase_ur}
\end{table*}%

\begin{table*}[t]\footnotesize
\centering
\caption{Parameter sensitivity analysis of the unlearning ratio (UR) on CIFAR100 dataset.}
\def\arraystretch{1.3}\begin{tabular}{c|ccc|ccc|c|c}
\toprule
 & \multicolumn{3}{c|}{\textbf{Consistency}}  & \multicolumn{3}{c|}{\textbf{Accuracy}} & \textbf{Efficiency} & \textbf{Verifiability}  \\  \midrule
UR  & $Con(\mathbf X^{r})$  & $Con(\mathbf X^{t})$  & $Con(\mathbf X^{u})$  & $Acc(\mathbf X^{r})$ & $Acc(\mathbf X^{t})$ & $Acc(\mathbf X^{u})$ & seconds & $|\Delta|$ \\  \midrule
0.1\% & $0.004 \pm 0.000$  & $0.126 \pm 0.003$  &  $0.228 \pm 0.008$ & $1.000 \pm 0.000$ & $0.783 \pm 0.002$ & $0.763 \pm 0.018$ & $960.33 \pm 3.33$ & $0.250^{***}$ \\ 
0.5\% & $0.004 \pm 0.000$  & $0.128 \pm 0.003$  &  $0.211 \pm 0.014$ & $1.000 \pm 0.000$ & $0.777 \pm 0.002$ & $0.781 \pm 0.023$ & $938.14 \pm 2.24$ & $0.223^{***}$ \\ 
1.0\% & $0.005 \pm 0.000$  & $0.130 \pm 0.003$  &  $0.204 \pm 0.009$ & $1.000 \pm 0.000$ & $0.776 \pm 0.003$ & $0.776 \pm 0.015$ & $958.23 \pm 0.49$ & $0.228^{***}$ \\ 
5.0\% & $0.007 \pm 0.001$  & $0.133 \pm 0.003$  &  $0.194 \pm 0.004$ & $1.000 \pm 0.000$ & $0.772 \pm 0.002$ & $0.837 \pm 0.006$ & $966.01 \pm 2.17$ & $0.197^{***}$ \\ 
10.0\% & $0.010 \pm 0.001$  & $0.141 \pm 0.008$  &  $0.198 \pm 0.008$ & $1.000 \pm 0.000$ & $0.764 \pm 0.004$ & $0.851 \pm 0.002$ & $977.92 \pm 1.83$ & $0.198^{***}$ \\ \bottomrule
 \multicolumn{3}{l}{\textit{Notes.} ***$p \leq 0.001$; **$p \leq 0.01$.}
\end{tabular}%
\label{tab:cifar_ur}
\end{table*}%

Table~\ref{tab:purchase_ur} and Table~\ref{tab:cifar_ur} present the performance of ETID with varying unlearning ratios (UR) on Purchase and CIFAR100, respectively. ETID exhibits stable performance across consistency, accuracy, efficiency, and verifiability with various unlearning ratio settings on both datasets. The results verify that our proposed framework ETID can effectively cope with unlearning tasks under different unlearning rates, maintaining high-performance standards across all aspects of machine unlearning desiderata. This robustness makes ETID a versatile and reliable choice for various unlearning scenarios, ensuring that companies can confidently implement unlearning protocols without sacrificing model performance or efficiency.

\section{Conclusion and Discussion}
In this study, we respond to the urgent calls for the RTBF as stipulated in various recent privacy regulations like GDPR and implement RTBF in data-driven predictive services. Our work proposes a holistic machine learning-to-unlearning framework ETID to handle the data erasure requests for predictive models, by integrating a novel ensemble modeling method and a new iterative information distillation method. Using datasets corresponding to two business predictive services, we demonstrate that ETID surpasses several state-of-the-art machine unlearning methods, comprehensively fulfilling the desiderata of machine unlearning across all aspects.


Our work makes several research contributions to the extant literature. First, to the best of our knowledge, we are one of the first to investigate the machine unlearning problem within the realm of business research. By addressing this problem from a systemic perspective, our work proposes the first holistic machine learning-to-unlearning framework. This framework includes innovative designs at both the predictive model construction stage and the unlearning requests response stage, in contrast to existing methods that typically focus on unlearning designs at only one stage. Thus, our study contributes to the current IS literature by introducing a novel method to the expanding repertoire of techniques that address critical business and societal issues related to privacy protection~\citep{smith11,xuanddinev22}. Second, in our proposed framework, we develop a novel ensemble learning method to build predictive models. On the one hand, it ensures that each sub-model is trained with most of the training data, avoiding the issue of low accuracy due to insufficient training. On the other hand, it establishes suitable reference models for the subsequent distillation-based unlearning process, facilitating the creation of accurate and consistent unlearned models. In addition, we have also designed a new distillation-based unlearning method to effectively and efficiently erase the information of unlearning samples from the target ensemble model. These innovative designs represent the methodological contributions of our study. Third, our framework is highly flexible and extensible. Specifically, our framework applies to various machine learning algorithms, allowing predictive service providers to use different algorithms to build their predictive models. Finally, we conduct extensive experiments with two datasets in predictive service scenarios. The experimental results demonstrate the superior performance of our framework over several state-of-the-art machine unlearning benchmarks. In particular, it not only constructs an ensemble predictive model with high accuracy performance but also can address the unlearning requests more efficiently and provide a more consistent and accurate unlearned model. Besides, the verifiability of the unlearned model generated by our method is also guaranteed.

This study also offers several managerial implications. First, it highlights the importance of adopting a holistic design when dealing with data erasure requests in predictive services to potentially fulfill machine unlearning desiderata. By integrating comprehensive designs, predictive service providers can ensure that data erasure processes are thorough and effective. Second, predictive service providers can leverage our proposed framework ETID to efficiently handle unlearning requests from data subjects, thereby mitigating the risk of fines for breaching the RTBF. According to GDPR regulations, this could potentially help companies save up to EUR 20 million or 4\% of the company's global turnover\footnote{https://europa.eu/youreurope/business/dealing-with-customers/data-protection/data-protection-gdpr/}. Additionally, by implementing ETID, companies can simultaneously preserve the accuracy of their predictive services, which is crucial for profitable operations. Taking Google as an example, considering that revenues from predictive analytics constitute an important part of Google’s \$305.63 billion total revenues\footnote{https://www.statista.com/statistics/267606/quarterly-revenue-of-google/}, such improvement could translate into significant financial gains for Google. This dual benefit of regulatory compliance and operational efficiency positions ETID as a valuable tool for predictive service providers. Third, our framework enhances data erasure capabilities for data subjects and potentially promotes the data-driven predictive service market. By providing robust mechanisms for data subjects to exercise their right to be forgotten, ETID fosters greater trust in data-driven services. This trust, in turn, can lead to increased adoption and growth of the predictive service market. Finally, beyond privacy assurance through data unlearning, our ETID framework holds promise for a range of additional applications. For instance, it can address discrimination issues arising from biased data in recruitment and credit scoring models, mitigate harmful effects on predictive models caused by misleading or malicious content, and remove the negative impacts of outdated data in fields with evolving information, such as finance and healthcare. By addressing these issues, ETID can bolster predictive models' fairness, security, and reliability. This broader applicability makes ETID a versatile tool for improving the overall quality and trustworthiness of predictive services. In conclusion, the adoption of ETID not only addresses immediate concerns related to data unlearning but also provides long-term benefits by enhancing model integrity and fostering a more trustworthy data ecosystem. This makes it a critical consideration for any organization or company involved in predictive analytics.

We further discuss the scenario in which a predictive model has been in use for an extended period, and a substantial amount of data has been forgotten. In this case, the condition described in Proposition~\ref{proposition2} (see E.q.~\ref{eq:prop2_condition}) may no longer be satisfied, potentially reducing the business value of the unlearned model. To address these issues, predictive service providers can incorporate new data into the model. Specifically, they can collect new data and periodically rebuild the predictive model using the ROEL method. Moreover, we suggest the distillation technique may offer a more efficient solution. In this approach, service providers first add new, non-overlapping data samples to each data partition pre-defined by ROEL, ensuring that these partitions remain of equal size. They can then construct new subsets using these partitions as ROEL did. Finally, the distillation technique can be applied to fine-tune the corresponding sub-models on the new data subsets, analogous to the model rectification process described in Section~\ref{sec:rectify}. Using these approaches, service providers can increase the amount of common data samples between sub-models, effectively yielding retrained-like reference models to guarantee the business value of the unlearned model.

There are some future research directions that merit attention. Firstly, we note that the implementation and good performance of ETID require some realistic assumptions and conditions. Specifically, ETID needs the unlearning data to accomplish the machine unlearning process. In this work, we assume that the unlearning data is available to predictive service providers during the unlearning process, as in the literature. This is reasonable since it's not prohibited by the current RTBF. However, if the RTBF in the future requires a "stricter forgetting" where the unlearning data cannot be used in any way once the unlearning request is initiated~\citep{tarun23}, it will become a very challenging problem. Thus, it would be interesting to study machine unlearning under this more stringent condition. Secondly, we mainly consider a realistic scenario where the size of unlearning data is not large. If the unlearning data is a very large subset of the training data (or even the entire training data), it would be extremely difficult to meet all the desiderata of machine unlearning, especially the business desiderata like accuracy. Therefore, studying the fine-grained impacts of the size of unlearning data on the desiderata of machine unlearning is also a very promising future research direction. Lastly, while ideally addressing data erasure requests for standard predictive models, this study illuminates some future research directions in machine unlearning, especially for the burgeoning Large Language Models (LLMs)~\citep{zhang23,liu24} and federated predictive modeling~\citep{gao24}.

\bibliographystyle{nonumber}

\clearpage

\section*{Appendices}

In this Appendix, we provide the proofs of the propositions in Section A; besides, we introduce the detailed experimental designs of verifiability examination in Section B.

\subsection*{A. Proofs}
\label{appendix:proofs}
\subsubsection*{A.1 Proof of Proposition~\ref{proposition1}}
\begin{proposition_appendix}
    Given the sub-models generated by ROEL and unlearning data $\mathbf D^u \subset \mathbf d_i$, sub-model $M_i$ is a retrained-alike model of sub-model $M_j, \forall j \neq i$.
\end{proposition_appendix}
\textbf{Proof.} We have sub-model $M_i=A(\mathbf D_{-i})=A(\mathbf D/\mathbf d_i)$; moreover, the na\"ive retrained model of sub-model $M_j$ is obtained by $M^{rt}_{j}=A(\mathbf D_{-j}/\mathbf D^u)=A(\mathbf D/(\mathbf d_j \cup \mathbf D^u))$. Let further denote $\mathbf D_{-ij}=\mathbf D / (\mathbf d_i \cup \mathbf d_j)$. Then, the sub-model $M_i$ can be rewritten as: $$M_i=A(\mathbf D_{-ij} \cup \mathbf d_j).$$
The na\"ive retrained model of sub-model $M_j$ can be rewritten as:
$$M^{rt}_j=A(\mathbf D_{-ij} \cup (\mathbf d_i / \mathbf D^u)).$$
Since $\mathbf d_i \cap \mathbf d_j=\emptyset$, for $\forall i \neq j$, and $\mathbf D^u \subset \mathbf d_i$, the shared training samples of $M_i$ and $M^{rt}_{j}$ are $\mathbf D_{-ij}$; furthermore, the unique training samples of $M_i$ and $M^{rt}_{j}$ are $\mathbf d_j$ and $\mathbf d_i/\mathbf D^u$, respectively. Thus, according to Definition~\ref{def_delta_alike}, we have:
$$\frac{|\mathbf D_{-ij}|}{\max (|\mathbf d_j|, |\mathbf d_i/\mathbf D^u|)}=\frac{|\mathbf D / (\mathbf d_i \cup \mathbf d_j)|}{\max (|\mathbf d_j|, |\mathbf d_i/\mathbf D^u|)}=\frac{|(\mathbf d_1 \cup \mathbf d_2 \cup \cdots \cup \mathbf d_K) / (\mathbf d_i \cup \mathbf d_j)|}{\max (|\mathbf d_j|, |\mathbf d_i/\mathbf D^u|)} \geq \delta$$
As $K\geq3$ and $\mathbf d_1,\mathbf d_2,\cdots,\mathbf d_K$ are of equal size, thus $\delta \geq 1$, \emph{i.e.}, sub-model $M_i$ is a $\delta$-alike ($\delta \geq 1$) model of $M^{rt}_{j}$; in another word, sub-model $M_i$ is a retrained-alike model of $M_{j}$.

\subsubsection*{A.2 Proof of Proposition~\ref{proposition2}}
\begin{proposition_appendix}
    Given the unlearned sub-models $\{M_{j}|j\in[1,K]\}$, updated reference models $\{M^{rf}_{i}|i\in[1,K]\}$, and new coming unlearning data $\mathbf D^{u}{'} \subset \mathbf d_i$, $M^{rf}_{i}$ can still be considered a retrained-alike model of sub-model $M_j, \forall j \neq i$, if and only if: $$\frac{|\mathbf D^{r}_{-i} \bigcap \mathbf D^{r}_{-j}|}{\max(|\mathbf D^{r}_{-i}|-|\mathbf D^{r}_{-i} \bigcap \mathbf D^{r}_{-j}|,|\mathbf D^{r}_{-j}|-|\mathbf D^{r}_{-i} \bigcap \mathbf D^{r}_{-j}|)}\geq 1.$$ where $\mathbf D^r_{-i}=\mathbf D_{-i} / \mathbf D^u$ and $\mathbf D^r_{-j}=\mathbf D_{-j} / \mathbf D^u$ denote the remaining data of sub-model $M_i$ and $M_j$, respectively.
\end{proposition_appendix}
\textbf{Proof.} The proof of Proposition~\ref{proposition2} is similar to that of Proposition~\ref{proposition1}. We can deduce that updated reference model $M^{rf}_{i}$ is a $\delta$-alike ($\delta \geq 1$) model of $M^{rt}_{j}$ (\emph{i.e.}, the na\"ive retrained version of $M_{j}$). Therefore, $M^{rf}_i$ is a retrained-alike model of $M_{j}$ according to Definition~\ref{def_retrained_alike}.

\subsubsection*{A.3 Proof of Proposition~\ref{proposition3}}
\begin{proposition_appendix}
    Given sub-model $M_j$ and its unlearned sub-model $M^u_j$ generated by ETID, $M^u_j$ is distinguishable from $M_j$, for $\forall j \in [1,K]$.
\end{proposition_appendix}
\textbf{Proof.} We have the sub-model $M_j=A(\mathbf D_{-j})=A(\mathbf D/ \mathbf d_j)$, and the sub-model $M_i=A(\mathbf D_{-i})=A(\mathbf D/ \mathbf d_i)$. Assuming the unlearning data $\mathbf D^u \subset \mathbf d_i$, according to Proposition~\ref{proposition1}, the sub-model $M_i$ can serve as a suitable reference model to refine $M_j$ to achieve unlearning by distillation.

Using ETID, the unlearned sub-model $M^u_j$ is generated under the supervision of $M_i$ with the objective Eq.\ref{eq:unlearning}, by enforcing the outputs of $M^u_j$ on unlearning data $\mathbf D^u$ to be close to those of $M_i$, \emph{i.e.}, $M^u_j(\mathbf X^u) = M_i(\mathbf X^u)$. As the outputs of $M_j$ on $\mathbf D^u$ are different from those of $M_i$, \emph{i.e.}, $M_j(\mathbf X^u) \neq M_i(\mathbf X^u)$, we can take the difference between the outputs of these two models, $M^u_j(\mathbf X^u)$ and $M_j(\mathbf X^u)$, as a verification function to distinguish them from each other.

\subsection*{B. Verifiability Examination}
\label{appendix:verifiability}
\begin{figure}[htbp]
\begin{center}
\includegraphics[width=0.8\linewidth]{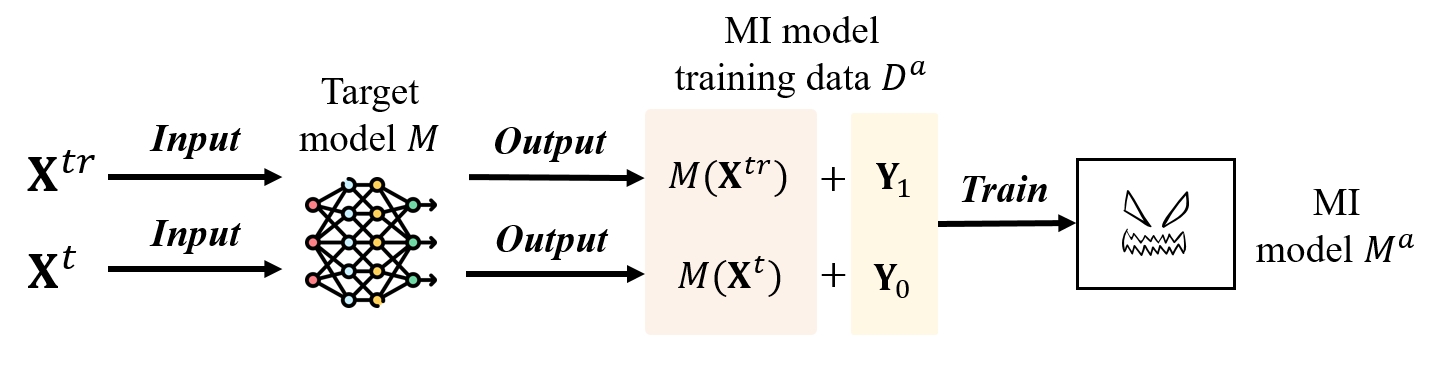}
\caption{MI model construction.}
\label{fig:mia_construction}
\end{center}
\end{figure}

\begin{figure}[htbp]
\begin{center}
\includegraphics[width=0.95\linewidth]{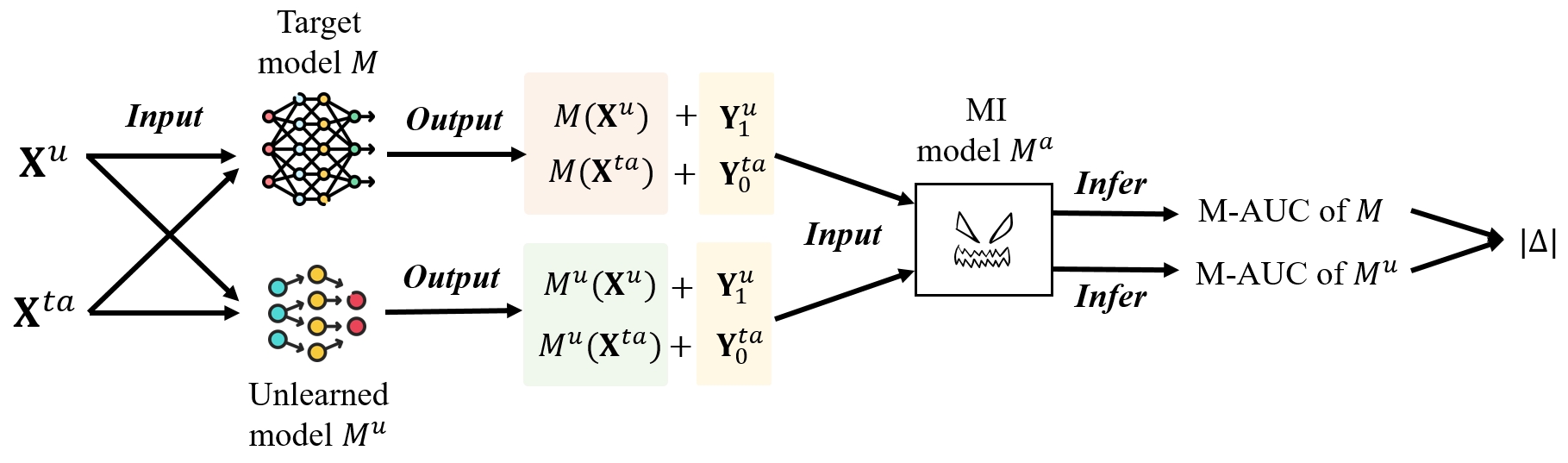}
\caption{M-AUC-based unlearning verifiability examination.}
\label{fig:mia_verification}
\end{center}
\end{figure}

In this part, we demonstrate how we use the membership inference to conduct the verifiability examination in detail. It includes two steps, Membership Inference (MI) model construction (as illustrated in Figure~\ref{fig:mia_construction}) and M-AUC-based examination (as shown in Figure~\ref{fig:mia_verification}). We first introduce some notations. Specifically, we use $\mathbf D^t$ to denote the testing dataset, $\mathbf X^t$ denotes the features of the testing samples; while $\mathbf D^{tr}$ is the same number of randomly sampled training samples as $\mathbf D^t$, $\mathbf X^{tr}$ is the features of these training samples. It is worth noting that $\mathbf D^{tr}$ and the unlearning data $\mathbf D^u$ are non-overlapping. Moreover, we use $\mathbf X^u$ to denote the features of the unlearning data. $\mathbf D^{ta}$ is the same number of randomly selected testing samples as $\mathbf D^u$, which is regarded as non-members to test the MI model, $\mathbf X^{ta}$ denoting their features. We further define $\mathbf Y_1$ and $\mathbf Y_0$ as the membership label vectors of member samples $\mathbf D^{tr}$ (with label "1") and non-member samples $\mathbf D^t$ (with label "0"), respectively.

In the first step, we input the features of training samples $\mathbf D^{tr}$ and the same number of testing samples $\mathbf D^{t}$ into the target model $M$, and obtain the corresponding outputs $M(\mathbf X^{tr})$ and $M(\mathbf X^{t})$, respectively. Then, we combine $M(\mathbf X^{tr})$ and $M(\mathbf X^{t})$ (as features), and $\mathbf Y_1$ and $\mathbf Y_0$ (as labels) to construct a training dataset $\mathbf D^a$ for MI model. Next, we train a binary MI model $M^a$ using a two-layer fully-connected neural network with the training dataset $\mathbf D^a$. During the examination step, we input the unlearning samples $\mathbf D^u$ and the same number of testing samples $\mathbf D^{ta}$ into the target model $M$ to obtain the outputs $M(\mathbf X^u)$ and $M(\mathbf X^{ta})$, and provide the membership label vector $\mathbf Y^u_1$ and $\mathbf Y^{ta}_0$, respectively; similarly, we obtain $M^u(\mathbf X^u)$ and $M^u(\mathbf X^{ta})$ by the unlearned model $M^u$, also company with $\mathbf Y^u_1$ and $\mathbf Y^{ta}_0$. Next, we respectively input the outputs of the target model $M$ and the unlearned model $M^u$ into MI model $M^a$, and calculate the M-AUC scores of two models. Finally, we obtain the absolute difference of the two M-AUC scores.


\end{document}